\documentclass[12pt]{elsarticle}
\usepackage[utf8]{inputenc}
\usepackage{setspace}
\usepackage[margin=0.75in]{geometry}
\usepackage{graphicx}
\graphicspath{ {./figures/} }
\usepackage[labelfont=bf, font = normalsize]{caption}
\usepackage{subcaption}
\usepackage{amsmath,amssymb,mathtools}
\usepackage{mathrsfs}
\usepackage{bm}
\usepackage{xcolor}
\usepackage{array}
\usepackage{appendix}
\usepackage{tikz}
\usepackage{floatrow}
\usepackage{rotating}
\usepackage{multirow,bigdelim}
\usepackage{array, blkarray, makecell}
\usepackage{tabularx}
\DeclareFloatFont{small}{\footnotesize}% "scriptsize" is defined by floatrow, "tiny" not
\newcolumntype{P}[1]{>{\centering\arraybackslash}p{#1}}

\providecommand{\keywords}[1]
{
  \small	
  \textbf{\textit{Keywords:}} #1
}

\usepackage{siunitx}
\usepackage{tikz-3dplot}
\usepackage{graphicx}
\usetikzlibrary{calc,backgrounds}
\usepackage{pgfplotstable}
\usepgfplotslibrary{groupplots}
\usetikzlibrary{intersections}
\usetikzlibrary{positioning}
\usepgfplotslibrary{fillbetween}

\definecolor{rwth1}{RGB}{0,84,159}      % RWTH-Blau
\definecolor{rwth2}{RGB}{142,186,229}   % RWTH-Hellblau
\definecolor{rwth3}{RGB}{0,97,101}      % Petrol 
\definecolor{rwth4}{RGB}{0,152,161}     % Türkis
\definecolor{rwth5}{RGB}{87,171,39}     % Grün
\definecolor{rwth6}{RGB}{189,205,0}     % Maigrün
\definecolor{rwth7}{RGB}{255,237,0}     % Gelb
\definecolor{rwth8}{RGB}{246,168,0}     % Orange
\definecolor{rwth9}{RGB}{227,0,102}     % Magenta
\definecolor{rwth10}{RGB}{204,7,30}     % Rot
\definecolor{rwth11}{RGB}{161,16,53}    % Bordeaux
\definecolor{rwth12}{RGB}{97,33,88}     % Violett
\definecolor{rwth13}{RGB}{122,111,172}  % Lila

\definecolor{rwthb1}{HTML}{e8f1fa}      % RWTH-Blau1
\definecolor{rwthb2}{HTML}{c7ddf2}      % RWTH-Blau2
\definecolor{rwthb3}{HTML}{8ebae5}      % RWTH-Blau3
\definecolor{rwthb4}{HTML}{407fb7}      % RWTH-Blau4
\definecolor{rwthb5}{HTML}{00549f}      % RWTH-Blau5

\definecolor{rwtho1}{HTML}{fff7ea}      % RWTH-Orange1
\definecolor{rwtho2}{HTML}{feeac9}      % RWTH-Orange2
\definecolor{rwtho3}{HTML}{fdd48f}      % RWTH-Orange3
\definecolor{rwtho4}{HTML}{fabe50}      % RWTH-Orange4
\definecolor{rwtho5}{HTML}{f6a800}      % RWTH-Orange5

\definecolor{lightblue}{RGB}{173,216,230}

\tikzstyle{dashpattern0} = [dash pattern = ]
\tikzstyle{dashpattern1} = [dash pattern = on 4.25pt off 0.75pt]
\tikzstyle{dashpattern2} = [dash pattern = on 1.5pt off 0.5pt]
\tikzstyle{dashpattern3} = [dash pattern = on 0.75pt off 0.4pt]
\tikzstyle{dashpattern4} = [dash pattern = on 3pt off 1pt on 1pt off 1pt]
\tikzstyle{dashpattern5} = [dash pattern = on 3.75pt off 0.5pt on 0.75pt off 0.5pt on 0.75pt off 0.5pt]
\tikzstyle{dashpattern6} = [dash pattern = on 3.25pt off 0.5pt on 0.75pt off 0.5pt on 0.75pt off 0.5pt on 0.75pt off 0.5pt]
\tikzstyle{dashpattern7} = [dash pattern = on 3.25pt off 0.5pt on 0.75pt off 0.5pt on 0.75pt off 0.5pt on 0.75pt off 0.5pt on 0.75pt off 0.5pt]
\tikzstyle{dashpattern8} = [line cap=round, dash pattern = on 3.25pt off 2.75pt]
\tikzstyle{dashpattern9} = [line cap=round, dash pattern = on 0.01pt off 2pt]
\tikzstyle{dashpattern10}= [line cap=round, dash pattern = on 3.25pt off 2pt on 0.01pt off 2pt]
\tikzstyle{dashpattern11}= [line cap=round, dash pattern = on 3.5pt off 1.75pt on 0.01pt off 1.75pt on 0.01pt off 1.75pt]
\tikzstyle{dashpattern12}= [line cap=round, dash pattern = on 3.5pt off 1.75pt on 0.01pt off 1.75pt on 0.01pt off 1.75pt on 0.01pt off 1.75pt]
\tikzstyle{dashpattern13}= [line cap=round, dash pattern = on 3.5pt off 1.75pt on 0.01pt off 1.75pt on 0.01pt off 1.75pt on 0.01pt off 1.75pt on 0.01pt off 1.75pt]

\newcommand{\tikzvdots}{%
  \begin{tikzpicture}[baseline=(current bounding box.center)]
    \fill (0,0) circle (2.0pt);
    \fill (0,-0.2) circle (2.0pt);
    \fill (0,-0.4) circle (2.0pt);
  \end{tikzpicture}%
}
\newcommand{\tightI}{\text{I}}
\newcommand{\tightII}{\text{I\hspace{-0.2mm}I}}
\newcommand{\tightIII}{\text{I\hspace{-0.2mm}I\hspace{-0.2mm}I}}
\newcommand{\tightIV}{\text{I\hspace{-0.2mm}V}}
            
\usepackage{tablefootnote}    
\usepackage{algorithm}
\usepackage{algpseudocode}

\usepackage{hyperref}

\usepackage{placeins}

\usepackage{amsthm}

\newtheorem{remark}{Remark}  
            
\usepackage{natbib}
\title{%
	\Large A generalized dual potential for inelastic Constitutive Artificial Neural Networks: \large A JAX implementation at finite strains}

\author[1]{Hagen Holthusen \corref{cor1}}
\ead{hagen.holthusen@ifam.rwth-aachen.de}
\cortext[cor1]{Corresponding author\\hagen.holthusen@ifam.rwth-aachen.de\\ Mies-van-der-Rohe-Str. 1, 52074 Aachen, Germany}
\author[1,2]{Kevin Linka}
\author[3]{Ellen Kuhl}
\author[1]{Tim Brepols}

\address[1]{Institute of Applied Mechanics, RWTH Aachen University, Germany}
\address[2]{Institute for Continuum and Material Mechanics, Hamburg University of Technology, Germany}
\address[3]{Department of Mechanical Engineering, Stanford University, United States}
\date{}

\begin{document}

%%%%%% Abstract %%%%%%

\begin{abstract}
We present a methodology for designing a generalized dual potential, or pseudo potential, for inelastic Constitutive Artificial Neural Networks (iCANNs). 
This potential, expressed in terms of stress invariants, inherently satisfies thermodynamic consistency for large deformations. 
In comparison to our previous work, the new potential captures a broader spectrum of material behaviors, including pressure-sensitive inelasticity.

To this end, we revisit the underlying thermodynamic framework of iCANNs for finite strain inelasticity and derive conditions for constructing a convex, zero-valued, and non-negative dual potential. 
To embed these principles in a neural network, we detail the architecture's design, ensuring a priori compliance with thermodynamics.

To evaluate the proposed architecture, we study its performance and limitations discovering visco-elastic material behavior, though the method is not limited to visco-elasticity.
In this context, we investigate different aspects in the strategy of discovering inelastic materials.
Our results indicate that the novel architecture robustly discovers interpretable models and parameters, while autonomously revealing the degree of inelasticity.

The iCANN framework, implemented in JAX, is publicly accessible at \url{https://doi.org/10.5281/zenodo.14894687}.\\
\end{abstract}

\maketitle

\section{Introduction}
\label{sec:introduction}
Constitutive material modeling plays a crucial role in understanding the behavior of history-dependent materials, as it allows for accurate predictions of their response under various loading conditions. The selection of an appropriate material model requires extensive expertise, particularly when addressing both elastic and inelastic behaviors which are often strongly coupled. A flawed choice can lead to a focus on optimizing material parameters rather than identifying the most suitable model for the material's behavior.

It is important to recognize that the expertise of professionals is often related to their practical experience, which can lead to biases in the modeling process. This subjectivity has the potential to influence the selection of models and parameters, thereby constraining the exploration of alternative approaches.

In this context, neural networks which are grounded in physical principles -- whether strongly or weakly -- can provide significant assistance, besides model-free approaches \cite{EggersmannKirchdoerferEtAl2019,PRUME2023115704}. Such neural networks are particularly valuable in the field of finite deformations, where complexity is a major challenge. By using advanced computational methods, we can improve our ability to capture the complex responses of materials under different conditions, ultimately leading to more reliable and effective engineering solutions and paving the way for novel advanced materials.
%
%============================================
\subsection{State-of-the-art}
\label{sec:state_art}
\textbf{Neural networks for inelastic materials.} In recent years, the use of neural networks has surged across various scientific fields, particularly for analyzing large data sets beyond human analytical capabilities. For instance, \citet{kepner2019} demonstrate the societal impact of the internet using hypersparse neural networks on a data set of 50 billion packets.

In medical science, neural networks significantly enhance the understanding of complex brain connectivity essential for identifying neurodegenerative diseases, as shown in \cite{xu2023datadrivennetworkneurosciencedata} with data from 2,702 subjects. Additionally, \citet{girardi2018patientriskassessmentwarning} utilize an attention-based neural network trained on 600,000 medical notes to detect critical warning symptoms.
Moreover, \citet{ferle2024predictingprogressioneventsmultiple} combine Long Short-Term Memory networks with Conditional Restricted Boltzmann Machines to predict multiple myeloma events up to twelve months in advance, improving patient care.
In the field of mechanics, \citet{Nagle2024} trained a neural network on 1,000 individual virtual subjects to predict skin growth.

In continuum mechanics, access to extensive experimental data is often limited, prompting the question of how neural networks can be beneficial in this field. Expert-based neural network architectures offer distinct advantages, particularly when modeling the diverse behaviors of inelastic materials. The phenomena of interest, such as visco-elasticity and plasticity, exhibit variability based on factors like pressure-sensitivity.

Typically, experts evaluate experimental data to determine both the inelastic phenomenon and the corresponding constitutive model equations. These decisions are often influenced by prior experience, introducing bias. Given the complexity of constitutive equations, it is challenging for humans to formulate highly nonlinear relationships accurately.

Neural networks address this limitation by accommodating a wide range of material behaviors and approximating complex nonlinear functions with high accuracy through scaling their depth and width. Recent research has focused on integrating neural networks with machine learning to enhance modeling capabilities in continuum mechanics.

One of the pioneering thermodynamics-based networks was introduced in \cite{masi2021}, focusing specifically on inelastic multiscale modeling \cite{masi2022} and the inelastic evolution process \cite{masi2023}. Further advancements by \cite{Piunno2025} employed proper orthogonal decomposition to extract macroscopic internal state variables, thereby enriching the data set for training the network.
A hierarchical discovery framework that aligns with thermodynamic principles has been proposed in \cite{ZHANG2025106049}.

\citet{asad2023cmame} proposed a mechanics-informed neural network \cite{asad2022} aimed at elucidating the behavior of visco-elastic solids. Physics-augmented neural networks \cite{Klein_Roth_Valizadeh_Weeger_2023,KALINA2024116739} exploit fundamental principles from continuum mechanics, such as polyconvexity of the Helmholtz free energy, facilitating discoveries related to visco-elasticity \cite{rosenkranz2024} and thermo-elasticity \cite{fuhg2024polyconvexneuralnetworkmodels}. In this context, \cite{TAC2022115248} proposed a methodology that integrates deformation invariants with the architecture of neural ordinary differential equations.

To uncover inelastic materials characterized by an inelastic potential, the idea of input convex neural networks \cite{Amos2017} is particularly helpful due to their ability to ensure thermodynamically consistent designs. For example, physics-informed neural networks developed for elasto-viscoplasticity discovery align with this network architecture \cite{EGHTESAD2024104072}.

The unsupervised learning framework EUCLID incorporates Generalized Standard Materials (GSM) -- a method also used by \cite{Flaschel2024convex} -- into its architecture to autonomously identify plasticity under small strain conditions \cite{flaschel2022,flaschel2023}. This framework has recently been extended to address non-associative pressure-sensitivity under small strains \cite{Xu2025}. Additionally, \cite{upadhyay2024} utilized a dissipation potential based on the rate of the right Cauchy-Green tensor to characterize finite visco-elastic behavior within a GSM framework.

In scenarios involving damage, a built-in physics neural network was proposed by \cite{TAC2024102220}, while fracture problems were addressed through generalizable symbolic regression techniques in \cite{YI2025105916}. Moreover, a deep neural network capable of automatically locating and inserting regularized discontinuities for modeling brittle fractures was introduced in \cite{Baek2024}.

We adopt the approach established by Constitutive Artificial Neural Networks (CANNs) \cite{LinkaHillgaertnerEtAl2021,LinkaKuhl2023}, which were extended to visco-elasticity based on a Prony series in \cite{AbdolaziziLinkaEtAl2023}.
Recently, the idea of CANNs served as foundation for constitutive Kolmogorov Arnold networks \cite{abdolazizi2025constitutivekolmogorovarnoldnetworksckans}.
Its generalization to inelastic behavior (iCANN) under finite strains was proposed in \cite{holthusen2023,holthusen2023PAMM} and investigated in studies on visco-elasticity and elasto-plasticity with kinematic hardening \cite{boes2024arxiv}, as well as applications to biological tissue growth \cite{holthusen2025growth}. Although this custom-designed architecture inherently satisfies thermodynamic consistency for inelastic materials, it has certain limitations regarding scalability concerning depth and width and fails to incorporate pressure sensitivity effectively. 

Therefore, in this study, we explore methodologies for transitioning traditional feed-forward architectures onto the iCANN framework. Our novel architecture has similarities to recent contributions from \citet{JADOON2025117653}, who pertain to finite elasto-plasticity and share conceptual foundations with iCANNs. However, unlike  \citet{JADOON2025117653}, we fully custom-design our feed-forward network based on comprehensive discussions aimed at achieving a thermodynamically consistent yet generalized potential.

This overview of neural networks applied within the realm of thermodynamically consistent discovery of inelastic materials is not exhaustive; indeed, data-driven mechanics is an expanding field. Therefore, interested readers are encouraged to consult recent survey articles such as \cite{linden2023,watson2024arxiv,Fuhg2024}, which discuss methodologies for integrating physics into neural network architectures.\newline

\textbf{Discovery of inelastic neural networks.} As mentioned by \cite{Battalgazy2025}, different constitutive models can describe specific mechanical behaviors depending on parameter variations. 
The authors proposed a Bayesian-based procedure that combines model selection with parameter identification. 
Similarly, neural networks in continuum mechanics face challenges; as their architectures become more generalized, encompassing various inelastic phenomena, the complexity of the discovery process increases.

While highly generalizable, dense networks offer numerous advantages, they pose challenges in uniquely identifying the network's weights.
In this context, obtaining a sparse network during the discovery process is preferred over a dense one, as it can be considered `unique' to a certain extent. 
Regularizing the network's weights in thermodynamics-based architectures has proven to be a valuable tool in this regard \cite{mcculloch2024}.
This emphasis on sparsity not only enhances the model’s interpretability but also aligns with the objective of extracting meaningful insights from the underlying physical processes.

This consideration becomes increasingly critical in scenarios where data is subject to uncertainties \cite{LINKA2025117517}, as establishing a deterministic relationship between stresses and strains may often be unrealistic. 
Strategically, incrementally increasing the network's complexity could facilitate the identification of a unique network configuration \cite{linka2024} .

The challenge of creating a `rich' data set that enables optimal training of neural networks is gaining importance. 
On a structural level, this raises questions about how to design test specimens that maximize information extraction, which has been explored for one-shot identification in \cite{ghouli2025topologyoptimisationframeworkdesign}. 
Additionally, microstructural simulations serve as powerful tools for enriching macroscopic data sets \cite{PRUME2025117525}. 
In this regard, the work presented in \cite{ZHANG2024112661} provides methodologies for reconstructing microstructures from extremely limited data sets.

%============================================
\subsection{Hypothesis}
\label{sec:hypothesis}
We hypothesize that leveraging the depth and width of neural networks within a dual potential framework enables the modeling and discovery of a broad spectrum of inelastic material behaviors, including pressure-sensitive inelastic flow. By scaling the complexity of the neural network architecture, we aim to capture increasingly sophisticated material responses. Furthermore, we anticipate that utilizing sufficiently `rich' data sets will lead to accurate model development and reliable training outcomes.

In this initial phase of our work, we simplify the problem by neglecting both intrinsic and induced directional influences. Consequently, we do not incorporate preferred directions (i.e.\ anisotropy) in the formulation of the Helmholtz free energy, we disregard hardening effects, and we assume that the potential depends solely on the driving force.

To ensure full accessibility of the source code and to facilitate reproducibility, we implement the entire framework and training scheme using JAX \cite{jax2018github}.
%
%============================================
\subsection{Outline}
\label{sec:outline}
We begin with a description of the constitutive framework for inelastic materials at finite strains, applicable to a wide range of materials, in Section~\ref{sec:constitutive}.
There, we formulate the dual/pseudo potential in terms of stress invariants and comment on its connection to Generalized Standard Materials in Section~\ref{sec:GSM}.
In Section~\ref{sec:network}, we present the novel neural network architecture of the generalized dual potential embedded in a recurrent context, discussing time discretization schemes and our regularization approach.
Therefore, in Section~\ref{sec:convex_potential}, we briefly discuss mathematical properties to obtain a potential that satisfies thermodynamics a priori.
In Section~\ref{sec:resembling}, we show how well-known models, such as the \textit{von Mises} or \textit{Drucker-Prager} models, are included in the architecture.
Afterwards, in Section~\ref{sec:results}, we train our network on artificial and experimentally obtained data and evaluate its performance.
Our results are critically assessed in  Section~\ref{sec:discussion}, including a discussion on currently observed limitations of the approach.
Finally, in Section~\ref{sec:conclusion}, we conclude on the proposed method and outline possible future investigations.
\section{Constitutive framework for finite strain inelasticity}
\label{sec:constitutive}
In this section, we briefly outline the underlying constitutive framework for general inelastic materials at finite strains, which are modeled using the multiplicative decomposition of the deformation gradient.
For this, we introduce two fundamental scalar-valued quantities: 
The Helmholtz free energy, $\psi$, as well as a dual potential, $\omega$.
All constitutively dependent quantities can be derived from these thermodynamic potentials.
This dual potential approach is certainly not the only method for modeling inelastic behavior. We will elucidate its close relationship to Generalized Standard Materials \cite{halphen1975}, another well-established framework for characterizing inelasticity, in Section~\ref{sec:GSM}.\newline

\textbf{Kinematics.} We employ the multiplicative decomposition of the deformation gradient, $\bm{F}=\bm{F}_e\bm{F}_i$, into an elastic part, $\bm{F}_e$, and an inelastic part, $\bm{F}_i$, cf. \cite{Eckart1948,Kroener1959,sidoroff1974,rodriguez1994}.
Both determinants of the individual parts are greater than zero.
Conceptually, we introduce an intermediate configuration, relative to which the elastic response is characterized.
Unfortunately, the multiplicative decomposition is non-unique, i.e.\ we may superimpose any rotation $\bm{F}=\bm{F}_e\bm{Q}^{\dagger^T}\bm{Q}^\dagger\bm{F}_i=:\bm{F}_e^\dagger\bm{F}_i^\dagger$ where $\bm{Q}^\dagger \in \mathrm{SO}(3)$ with $\mathrm{SO}(3)$ denoting the special orthogonal group.
By employing the singular value decomposition, we recognize that $\bm{F}_i$ and $\bm{F}_i^\dagger$ share the same singular values, and thus, the same stretch tensor $\bm{U}_i$ resulting from the polar decomposition $\bm{F}_i=\bm{R}_i\bm{U}_i$ with $\bm{R}_i \in \mathrm{SO}(3)$.
Thus, we find $\bm{U}_i$ to be unique, and further, $\bm{F}_i^\dagger=\bm{R}_i^\dagger\bm{U}_i$ where $\bm{R}_i^\dagger=\bm{Q}^\dagger\bm{R}_i$.
Lastly, we introduce an appropriate stretch measure of the elastic stretches $\bm{C}_e = \bm{F}_e^T\bm{F}_e = \bm{Q}^\dagger\bm{F}_e^{\dagger^T}\bm{F}_e^\dagger\bm{Q}^{\dagger^T}$, which however, is non-unique.\newline

\textbf{Clausius-Planck inequality.} Any constitutive framework for solids must satisfy the Clausius-Planck inequality $\mathcal{D} := -\dot{\psi} + 1/2\, \bm{S}:\dot{\bm{C}} \geq 0$ where $\bm{S}$ denotes the second Piola-Kirchhoff stress, while $\bm{C}=\bm{F}^T\bm{F}$ refers to the right Cauchy-Green tensor.
For the time being, we assume the Helmholtz free energy to be a scalar-valued isotropic function \cite{spencer1971,zheng1994} depending solely on $\bm{C}_e$, i.e. $\psi = \hat{\psi}(\bm{C}_e)$.
Hence, we obtain the following, cf. \cite{dettmer2004}
\begin{equation}
\mathcal{D} =    \left(\bm{S} - 2\,\bm{F}_i^{-1}\,\frac{\partial\psi}{\partial\bm{C}_e}\,\bm{F}_i^{-T} \right) : \frac{1}{2}\,\dot{\bm{C}} + \underbrace{2\,\bm{C}_e\,\frac{\partial\psi}{\partial\bm{C}_e}}_{=: \bm{\Sigma}} : \underbrace{\dot{\bm{F}}_i\bm{F}_i^{-1}}_{=: \bm{L}_i} \geq 0
\label{eq:dissipation}
\end{equation}
where we introduce the elastic Mandel-like stress $\bm{\Sigma}$, which is symmetric since $\psi$ is an isotropic function of $\bm{C}_e$, cf. \cite{svendsen2001}\footnote{In the case of initial anisotropy, $\psi$ is usually assumed to be an isotropic function of $\bm{C}_e$ and a structural tensor $\bm{H}$. Note that in this case $\bm{\Sigma}$ is no longer symmetric.}.
Noteworthy, since $\bm{\Sigma}$ solely depends on $\bm{C}_e$, the elastic Mandel-like stress is also non-unique, i.e. $\bm{\Sigma}=\bm{Q}^{\dagger}\bm{\Sigma}^\dagger\bm{Q}^{\dagger^T}$.
Following the arguments of \cite{coleman1961,coleman1963,coleman1967}, we assume the term in brackets in Inequality~\eqref{eq:dissipation} to be zero, revealing the state law for $\bm{S}$.
Consequently, as $\bm{\Sigma}$ is symmetric, we can reduce the dissipation inequality to
\begin{equation}
    \mathcal{D}_{red} := \bm{\Sigma} : \bm{D}_i \geq 0
    \label{eq:dissipation_red}
\end{equation}
where $\bm{D}_i := \mathrm{sym}(\bm{L}_i)$ is the symmetric part of $\bm{L}_i$.
To satisfy the reduced dissipation inequality for arbitrary processes, we will introduce a dual potential, $\omega=\hat{\omega}(\bm{\Sigma})$, which is assumed to be a scalar-valued isotropic function in order to be independent of the superimposed rotation $\bm{Q}^\dagger$.\newline

\textbf{Co-rotated intermediate configuration.} We have observed that the relevant constitutive quantities, such as $\bm{C}_e$ and $\bm{\Sigma}$, suffer from an inherent rotational non-uniqueness.
This poses challenges in computing these quantities and derivatives with respect to those, for instance $\frac{\partial\psi}{\partial\bm{C}_e}$.
To address this issue in our numerical implementation, we adopt the approach suggested by \cite{holthusen2023} and introduce a co-rotated intermediate configuration.
In short, this approach pulls all non-unique quantities back by either $\bm{R}_i$ or $\bm{R}_i^\dagger$, i.e. $\bar{(\bullet)}:=\bm{R}_i^T\,(\bullet)\,\bm{R}_i = \bm{R}_i^{\dagger^T}\,(\bullet)^\dagger\,\bm{R}_i^\dagger$.
Consequently, we obtain the following unique quantities
\begin{equation}
    \label{tb_eq:10}
    \bar{\bm{C}}_e = \bm{U}_i^{-1}\bm{C}\bm{U}_i^{-1}, \quad \bm{S} = 2\,\bm{U}_i^{-1}\,\frac{\partial\psi}{\partial\bar{\bm{C}}_e}\,\bm{U}_i^{-1}, \quad \bar{\bm{\Sigma}} = 2\,\bar{\bm{C}}_e\,\frac{\partial\psi}{\partial\bar{\bm{C}}_e}, \quad \bar{\bm{D}}_i = \mathrm{sym}\left(\dot{\bm{U}}_i\,\bm{U}_i^{-1} \right).
\end{equation}
Noteworthy, the co-rotated pullback preserves both the symmetry as well as the eigenvalues, which is considered an advantage.\newline

\textbf{Potential-based evolution equation.} It remains to introduce an evolution equation for $\bar{\bm{D}}_i$ in a thermodynamic consistent way such that Inequality~\eqref{eq:dissipation_red} is satisfied for arbitrary processes.
Therefore, we postulate the existence of a pseudo potential \cite{kerstin1969}, which we may identify as the dual potential resulting from the Legendre-Fenchel transformation of the `classical' dissipation potential (see Section~\ref{sec:GSM}) known from Generalized Standard Materials \cite{halphen1975}, viz.\
\begin{equation}
    \bar{\bm{D}}_i = \frac{\partial \omega\left(\bar{\bm{\Sigma}}\right)}{\partial\bar{\bm{\Sigma}}}.
\label{eq:EvolutionEquation}
\end{equation}
According to \citet{germain1983}, the dissipation inequality is naturally fulfilled if $\omega$ is \textit{convex}, \textit{zero-valued}, and \textit{non-negative} with respect to $\bar{\bm{\Sigma}}$ \cite{holthusen2025growth}\footnote{We may understand the evolution equation as a subderivative $\bar{\bm{D}}_i \in \partial_{\bar{\bm{\Sigma}}}\omega\left(\bar{\bm{\Sigma}}\right)$ in case of non-smooth potentials, where $\partial_{\bar{\bm{\Sigma}}}\omega\left(\bar{\bm{\Sigma}}\right)$ denotes the subderivative with respect to $\bar{\bm{\Sigma}}$, see \cite{germain1998}.}.\newline

\textbf{Invariant representation.} As discussed above, the dual potential is assumed to be an isotropic function of $\bar{\bm{\Sigma}}$, and can, thus, be expressed in terms of its invariants.
Here, we choose the common stress invariants $I_1^{\bar{\bm{\Sigma}}}:=\mathrm{tr}(\bar{\bm{\Sigma}})$, $J_2^{\bar{\bm{\Sigma}}}:=1/2\,\mathrm{tr}(\mathrm{dev}(\bar{\bm{\Sigma}})^2)$, and $J_3^{\bar{\bm{\Sigma}}}:=1/3\,\mathrm{tr}(\mathrm{dev}(\bar{\bm{\Sigma}})^3)$.
With these invariants at hand, the evolution equation reduces to
\begin{equation}
    \bar{\bm{D}}_i = \frac{\partial\omega^*}{\partial\bar{\bm{\Sigma}}} = \frac{\partial\omega^*}{\partial I_1^{\bar{\bm{\Sigma}}}}\,\bm{I} + \frac{\partial\omega^*}{\partial J_2^{\bar{\bm{\Sigma}}}}\,\mathrm{dev}\left(\bar{\bm{\Sigma}}\right) + \frac{\partial\omega^*}{\partial J_3^{\bar{\bm{\Sigma}}}}\,\mathrm{dev}\left(\mathrm{dev}\left(\bar{\bm{\Sigma}}\right)^2\right)
\label{eq:EvolutionEquationInvars}
\end{equation}
where $\omega^*=\hat{\omega}^*\left(I_1^{\bar{\bm{\Sigma}}},\sqrt{J_2^{\bar{\bm{\Sigma}}}},\sqrt[3]{J_3^{\bar{\bm{\Sigma}}}}\right)$.
The square and cubic roots are calculated to ensure that all invariants share the same unit.
If we plug \eqref{eq:EvolutionEquationInvars} into the co-rotated version of Equation~\eqref{eq:dissipation_red}
\begin{equation}
    %\mathcal{D}_{red} := \frac{\partial\omega^*}{\partial I_1^{\bar{\bm{\Sigma}}}}\,I_1^{\bar{\bm{\Sigma}}} + \frac{\partial\omega^*}{\partial \sqrt{J_2^{\bar{\bm{\Sigma}}}}}\,\sqrt{J_2^{\bar{\bm{\Sigma}}}} + \frac{\partial\omega^*}{\partial \sqrt[3]{J_3^{\bar{\bm{\Sigma}}}}}\,\sqrt[3]{J_3^{\bar{\bm{\Sigma}}}} \geq 0,
    \mathcal{D}_{red} = \nabla\omega^*(\mathbf{z}) \cdot \mathbf{z} \geq 0, \quad \mathbf{z} = \begin{pmatrix}
        I_1^{\bar{\bm{\Sigma}}} \\
        \sqrt{J_2^{\bar{\bm{\Sigma}}}} \\
        \sqrt[3]{J_3^{\bar{\bm{\Sigma}}}}
    \end{pmatrix}
    \label{eq:reduced_dissipation_vector}
\end{equation}
we observe that the inequality is satisfied if $\omega^*$ is convex, zero-valued, and non-negative with respect to its arguments; however, this does not guarantee its convexity with respect to $\bar{\bm{\Sigma}}$, cf. \cite{collins2002}.
The reason for this lies in the indefinite Hessian of $J_3^{\bar{\bm{\Sigma}}}$ with respect to the Mandel-like stress.
Nevertheless, non-convex yield surfaces, which are typically modelled as a potential subtracted by a threshold such as the yield stress, are not only of significant practical relevance \cite{gluege2017,matzenmiller1995,baghous2022} but also amenable to numerical treatment \cite{sheng2011,pedroso2008}.
As $\omega^*$ includes the special case of being convex with respect to $\bar{\bm{\Sigma}}$, we consider this framework advantageous. 
%
%=================================
\subsection{Relation to Generalized Standard Materials for solids}
\label{sec:GSM}
In the following, we will explain the intrinsic relationship between the present modeling framework for iCANNs and the classical framework of Generalized Standard Materials, the latter of which is well-known in the literature (see e.g.\ \cite{halphen1975,germain1983,flaschel2023}). We start again with the Helmholtz free energy and assume, for the same reasons as explained above, that it is a scalar-valued isotropic function of quantities in the co-rotated intermediate configuration. Specifically, we assume a dependence of $\psi$ on $\bar{\bm{C}}_e$ only, i.e.\ $\psi = \hat{\psi}(\bar{\bm{C}}_e)$. More general cases in which $\psi$ additionally depends on further internal state variables or structural tensors are, of course, possible. However, since this does not lead to additional insights in the presentation that follows, we will not consider this case for simplicity.

Exploiting again the Clausius-Planck inequality and the chain rule of differentiation, we may arrive at
\begin{equation}
\label{tb_eq:1}
\mathcal{D} = -\dot{\psi} + \frac{1}{2}\, \bm{S}:\dot{\bm{C}} \geq 0 \qquad \Rightarrow \qquad \mathcal{D} = -2\,\frac{\partial \psi}{\partial \bar{\bm{C}}_e} : \frac{1}{2}\,\dot{\bar{\bm{C}}}_e + \bm{S}:\frac{1}{2}\,\dot{\bm{C}} \geq 0.
\end{equation}
Considering $\bar{\bm{C}}_e = \bm{U}_i^{-1}\,\bm{C}\,\bm{U}_i^{-1}$, the relation $\dot{\bm{C}} = \dot{\bm{U}}_i\,\bar{\bm{C}}_e\,\bm{U}_i + \bm{U}_i\,\dot{\bar{\bm{C}}}_e\,\bm{U}_i + \bm{U}_i\,\bar{\bm{C}}_e\,\dot{\bm{U}}_i$, and well-known properties of the scalar product of two second-order tensors, Inequality~\eqref{tb_eq:1} can directly be rewritten as
\begin{equation}
\label{tb_eq:2}
\mathcal{D} = \left(\bar{\bm{S}} - 2\,\frac{\partial \psi}{\partial \bar{\bm{C}}_e}\right):\frac{1}{2}\,\dot{\bar{\bm{C}}}_e + \bar{\bm{\Sigma}}:\bar{\bm{L}}_i \geq 0.
\end{equation}
Here, $\bar{\bm{S}} := \bm{U}_i\,\bm{S}\,\bm{U}_i$ is the second Piola–Kirchhoff stress tensor relative to the co-rotated intermediate configuration and $\bar{\bm{\Sigma}} = \bar{\bm{C}}_e\,\bar{\bm{S}}$ denotes the (up to this point generally unsymmetric) Mandel-like stress tensor in the very same configuration. With $\bar{\bm{L}}_i = \mathrm{sym}(\bar{\bm{L}}_i) + \mathrm{skew}(\bar{\bm{L}}_i) = \bar{\bm{D}}_i + \bar{\bm{W}}_i$, Expression~\eqref{tb_eq:2} is finally rewritten as
\begin{equation}
\label{tb_eq:3}
\mathcal{D} = \bar{\bm{S}}^{\,\mathrm{dis}}:\frac{1}{2}\,\dot{\bar{\bm{C}}}_e + \mathrm{sym}(\bar{\bm{\Sigma}}):\bar{\bm{D}}_i + \mathrm{skew}(\bar{\bm{\Sigma}}):\bar{\bm{W}}_i \geq 0
\end{equation}
where $\bar{\bm{S}}^{\,\mathrm{dis}} := \left(\bar{\bm{S}} - 2\,\frac{\partial \psi}{\partial \bar{\bm{C}}_e}\right)$ can be considered the irreversible or dissipative part of the stress $\bar{\bm{S}}$.

To fulfill dissipation inequality~\eqref{tb_eq:3}, it is now customary in the framework of Generalized Standard Materials to assume a scalar-valued dissipation potential\footnote{As in case of the Helmholtz free energy $\psi$, we may directly formulate the dissipation potential $\Omega$ as a scalar-valued isotropic function of quantities in the co-rotated intermediate configuration, in order to avoid any ambiguities due to arbitrary rotations of the intermediate configuration.} $\Omega = \hat{\Omega}\left(\dot{\bar{\bm{C}}}_e,\bar{\bm{D}}_i,\bar{\bm{W}}_i\right)$, expressed in terms of the strain-like rate quantities $\dot{\bar{\bm{C}}}_e$, $\bar{\bm{D}}_i$, and $\bar{\bm{W}}_i$, which is convex, non-negative, and zero-valued at the origin, i.e.\ $\hat{\Omega}(\bm{0},\bm{0},\bm{0}) = 0$. This potential is conveniently used to derive complementary laws for the thermodynamic conjugate forces, i.e.\footnote{In case of a non-smooth dissipation potential $\Omega$, the partial derivatives in \eqref{tb_eq:4} and the following should be understood as subderivatives.}:
\begin{equation}
\label{tb_eq:4}
\bar{\bm{S}}^{\,\mathrm{dis}} = \frac{\partial \Omega}{\partial \dot{\bar{\bm{C}}}_e}, \qquad \mathrm{sym}(\bar{\bm{\Sigma}}) = \frac{\partial \Omega}{\partial \bar{\bm{D}}_i}, \qquad \mathrm{skew}(\bar{\bm{\Sigma}}) = \frac{\partial \Omega}{\partial \bar{\bm{W}}_i}.
\end{equation}
As can be shown, with the above definitions, thermodynamic consistency of the formulation is naturally ensured.\newline

\textbf{Strain-rate independent dissipation potential.} Further consequences for the dissipation potential arise when dealing with solid materials, for which it is usually assumed that purely elastic, but otherwise arbitrary deformations ($\dot{\bar{\bm{C}}}_e \ne \bm{0}$, $\bar{\bm{D}}_i = \bar{\bm{W}}_i = \bm{0}$) do not cause any dissipation. In this case, it can be inferred from Inequality \eqref{tb_eq:3} that
\begin{equation}
\label{tb_eq:5}
\mathcal{D} = \bar{\bm{S}}^{\,\mathrm{dis}}:\frac{1}{2}\,\dot{\bar{\bm{C}}}_e = 0 \qquad \Rightarrow \qquad \bar{\bm{S}}^{\,\mathrm{dis}} = \frac{\partial \hat{\Omega}}{\partial \dot{\bar{\bm{C}}}_e} = \bm{0}.
\end{equation}
In other words, it can be concluded that the dissipation potential cannot be a function of the elastic strain rate $\dot{\bar{\bm{C}}}_e$ in this case\footnote{This does by no means preclude any strain-rate-dependent behavior of the material model, as the rate dependency can be reflected in the evolution of the internal state variables.}. Furthermore, as the second Piola-Kirchoff stress tensor relative to the intermediate configuration becomes $\bar{\bm{S}} = 2\,\frac{\partial \psi}{\partial \bar{\bm{C}}_e}$, this immediately leads to a symmetric Mandel-like stress tensor $\bar{\bm{\Sigma}} = \bar{\bm{C}}_e\,\bar{\bm{S}} =  2\,\bar{\bm{C}}_e\,\frac{\partial\psi}{\partial\bar{\bm{C}}_e}$ (see also \eqref{tb_eq:10}$_3$), i.e.\ $ \mathrm{sym}(\bar{\bm{\Sigma}}) = \bar{\bm{\Sigma}}$ and $\mathrm{skew}(\bar{\bm{\Sigma}}) = \bm{0}$. The Clausius-Planck inequality therefore reduces to 
\begin{equation}
\label{tb_eq:6}
\mathcal{D}_{red} = \bar{\bm{\Sigma}}:\bar{\bm{D}}_i\geq 0,
\end{equation}
such that the dissipation potential needs to be a function of $\bar{\bm{D}}_i$ only, i.e.\ $\Omega = \hat{\Omega}(\bar{\bm{D}}_i)$ with $\hat{\Omega}(\bm{0}) = 0$. The complementary law for $\bar{\bm{\Sigma}}$ is then obtained as
\begin{equation}
\label{tb_eq:11}
\bar{\bm{\Sigma}} = \frac{\partial \Omega}{\partial \bar{\bm{D}}_i}.
\end{equation}

\textbf{Dual dissipation potential in terms of stress-like quantities.}
Finally, by means of a Legendre-Fenchel transformation of $\Omega$, a dual dissipation potential $\omega$ in terms of the stress-like quantity $\bar{\bm{\Sigma}}$ can be derived:
\begin{equation}
\label{tb_eq:8}
    \omega = \hat{\omega}(\bar{\bm{\Sigma}}) = \sup_{\bar{\bm{D}}_i}\left(\bar{\bm{\Sigma}} : \bar{\bm{D}}_i - \hat{\Omega}\!\left(\bar{\bm{D}}_i\right)\right).
\end{equation}
The latter potential is also convex, non-negative, and zero-valued at the origin, i.e.\ $\hat{\omega}(\bm{0}) = 0$. It can be employed to define a thermodynamically consistent evolution equation (or complementary law) for the strain-like internal state variable $\bar{\bm{D}}_i$ via
\begin{equation}
\label{tb_eq:11}
    \bar{\bm{D}}_i = \frac{\partial \omega}{\partial \bar{\bm{\Sigma}}}.
\end{equation}

\textbf{Relation to iCANN framework.} The above shows that iCANNs perfectly fit into the framework of Generalized Standard Materials. The only difference to the more classical approach is that, in the iCANN modeling framework presented here, the dual dissipation potential \eqref{tb_eq:8} in terms of the stress-like variable $\bar{\bm{\Sigma}}$ is constructed and identified \emph{directly}. However, this alternative procedure is by no means unusual or disadvantageous, and has been proposed as an equally valid approach by other authors in the past (among many others, \cite{LemaitreChaboche1994,Chaboche1997,LeuschnerFritzenEtAl2015,HoltzmanChrysochoosEtAl2018}). One potential can always be constructed from the other, owing to the remarkable duality of the Legendre-Fenchel transformation. 
\section{Architecture of generalized iCANN}
\label{sec:network}
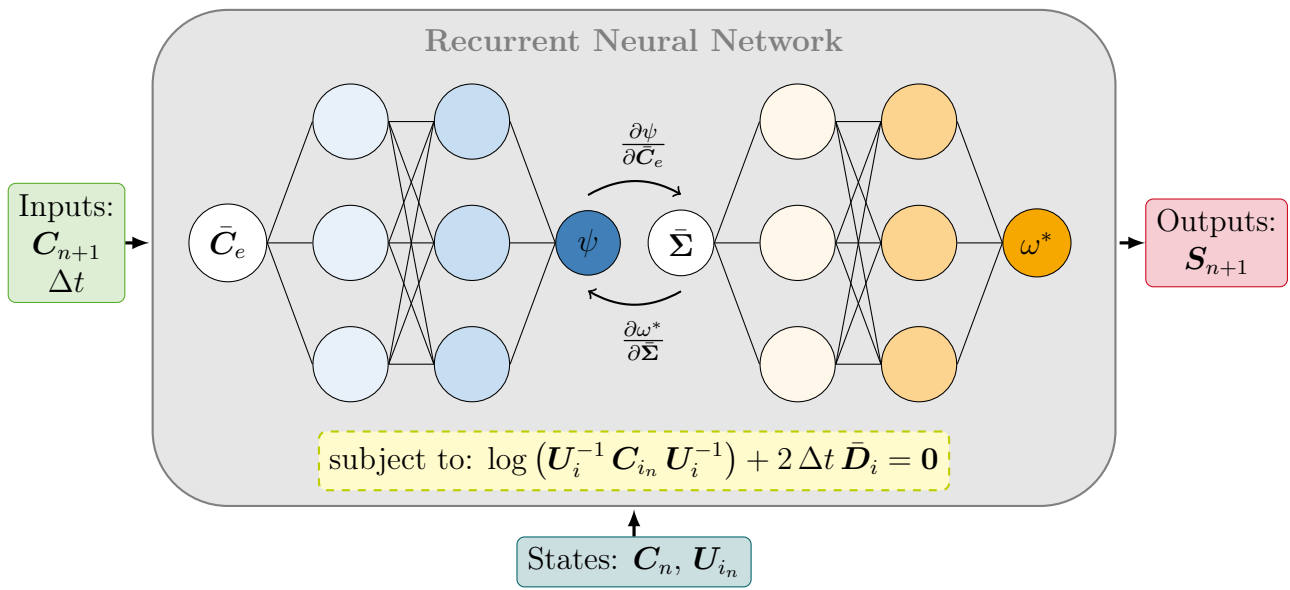
\begin{figure}[h]
    \centering
    \begin{tikzpicture}[node distance=0.6cm]
% rectangle
\fill[gray!20, rounded corners=1cm] (-1,-3.5) rectangle (11.8,3.1);
\draw[thick, gray, rounded corners=1cm] (-1,-3.5) rectangle (11.8,3.1);
\node (rl) at (-1, 0.0) {};
\node (rr) at (11.8, 0.0) {};
\node (rb) at (5.4, -3.4) {};
\node (rt) at (5.4, 2.7) [thick] {\textcolor{gray}{\textbf{Recurrent Neural Network}}};

% Subject
\node (subj) at (5.4, -2.9) [thick, dashed, draw=rwth6, rounded corners=0.1cm, fill=rwth7!20] {subject to: $\mathrm{log}\left(\bm{U}_i^{-1}\,\bm{C}_{i_n}\,\bm{U}_i^{-1}\right) + 2\,\Delta t\, \bar{\bm{D}}_i = \bm{0}$};

% State variables
\node[draw=rwth3, rounded corners=0.1cm, fill=rwth3!20] (states) at ($(rb.south) + (0,-0.7)$) {States: $\bm{C}_n,\,\bm{U}_{i_n}$};
\draw[->, very thick, >=latex] (states.north) -- (rb.south);

% Inputs
\node[align=center, draw=rwth5, rounded corners=0.1cm, fill=rwth5!20] (input) at ($(rl.west) + (-1,0)$) {Inputs: \\ $\bm{C}_{n+1}$ \\ $\Delta t$};
\draw[->, very thick, >=latex] (input.east) -- ($(rl.west)+(0.1,0)$);

% Outputs
\node[align=center, draw=rwth10, rounded corners=0.1cm, fill=rwth10!20] (output) at ($(rr.east) + (1.2,0)$) {Outputs: \\ $\bm{S}_{n+1}$};
\draw[->, very thick, >=latex] ($(rr.west)+(0.2,0)$) -- (output.west);

% Erstes Netzwerk
\node (input1) at ($(rl)+(1,0)$) [draw, circle, fill=white] {$\bar{\bm{C}}_e$};
\node (hidden1_1) [draw, circle, right=of input1, minimum size=1cm, fill=rwthb1] {};
\node (hidden1_2) [draw, circle, above=of hidden1_1, minimum size=1cm, fill=rwthb1] {};
\node (hidden1_3) [draw, circle, below=of hidden1_1, minimum size=1cm, fill=rwthb1] {};
\node (hidden2_1) [draw, circle, right=of hidden1_1, minimum size=1cm, fill=rwthb2] {};
\node (hidden2_2) [draw, circle, right=of hidden1_2, minimum size=1cm, fill=rwthb2] {};
\node (hidden2_3) [draw, circle, right=of hidden1_3, minimum size=1cm, fill=rwthb2] {};
\node (output1) [draw, circle, right=of hidden2_1, fill=rwthb4] {$\psi$};
%
%% Verbindungen erstes Netzwerk
\foreach \i in {1,...,3} {
    \draw[-] (input1.east) to (hidden1_\i.west);
}
\foreach \j in {1,...,3} {
	\foreach \k in {1,...,3} {
    		\draw[-] (hidden1_\j.east) to (hidden2_\k.west);
    	}
}
\foreach \k in {1,...,3} {
    \draw[-] (hidden2_\k.east) to (output1.west); 
}

% Zweites Netzwerk
\node (a) at ($(output1.east)+(0.2,0)$){};

\node (input2) [draw,circle,right of=a, fill=white]{$\bar{\bm{\Sigma}}$};
\node (hidden3_1) [draw, circle, right=of input2, minimum size=1cm, fill=rwtho1] {};
\node (hidden3_2) [draw, circle, above=of hidden3_1, minimum size=1cm, fill=rwtho1] {};
\node (hidden3_3) [draw, circle, below=of hidden3_1, minimum size=1cm, fill=rwtho1] {};
\node (hidden4_1) [draw, circle, right=of hidden3_1, minimum size=1cm, fill=rwtho3] {};
\node (hidden4_2) [draw, circle, right=of hidden3_2, minimum size=1cm, fill=rwtho3] {};
\node (hidden4_3) [draw, circle, right=of hidden3_3, minimum size=1cm, fill=rwtho3] {};
\node (output2) [draw, circle, right=of hidden4_1, fill=rwtho5] {$\omega^*$};
%
%% Verbindungen zweites Netzwerk
\foreach \i in {1,...,3} {
    \draw[-] (input2.east) to (hidden3_\i.west);
}
\foreach \j in {1,...,3} {
	\foreach \k in {1,...,3} {
    		\draw[-] (hidden3_\j.east) to (hidden4_\k.west);
    	}
}
\foreach \k in {1,...,3} {
    \draw[-] (hidden4_\k.east) to (output2.west); 
}
%

% connect
\path[->,thick] ($(output1.north) + (0,0.2)$) edge[bend left]($(input2.north) + (0,0.2)$);
\path[->,thick] ($(input2.south) + (0,-0.2)$) edge[bend left]($(output1.south) + (0,-0.2)$);

% derivatives
\node (deriv1) at ($(output1.east) + (0.3,1.3)$) {$\frac{\partial\psi}{\partial\bar{\bm{C}}_e}$};
\node (deriv2) at ($(output1.east) + (0.3,-1.3)$) {$\frac{\partial\omega^*}{\partial\bar{\bm{\Sigma}}}$};

\end{tikzpicture}
    \caption{Schematic illustrating two feed-forward networks (blue and orange) integrated within a recurrent architecture. The evolution equation for $\bm{U}_i$  can be solved either explicitly or implicitly. In the explicit approach, the state variable $\bm{C}_n$ is needed alongside $\bm{U}_{i_n}$. Conversely, the implicit integration method requires satisfying the evolution equation (equality constraint) but does not involve $\bm{C}_n$. The current stretch and time increment inputs $\bm{C}_{n+1}$ and $\Delta t$ yield the current stress $\bm{S}_{n+1}$, while the state variables are propagated through time.}
    \label{fig:RNN}
\end{figure}

\textbf{Recurrent architecture.} History-dependent materials require sequential data processing. 
The iCANN framework integrates two independent feed-forward networks -- one for dual potential discovery and another for the Helmholtz free energy -- within a recurrent architecture. Although distinct, these networks are strongly coupled via their derivatives, governing inelastic stretch and thermodynamic driving force. 
Their architectures are detailed in Sections~\ref{sec:FFN_potential} and \ref{sec:FFN_energy}.
Schematic~\ref{fig:RNN} depicts the overall structure. 
As typical in recurrent networks, the current output, $\bm{S}$, depends not only on the current input, $\bm{C}$, but also on propagated states, $\bm{U}_i$, across sequential time data. 
The time integration scheme, whether implicit or explicit, relies on the time increment $\Delta t$. Explicit methods require propagating the previous step’s right Cauchy-Green tensor, $\bm{C}_n$, while implicit methods solve for inelastic stretch iteratively via an equality constraint.\newline

\textbf{Loss function.} Defining the loss function is critical for accurate neural network training. While neural networks effectively learn complex input-output relationships, their flexibility can lead to overfitting. To mitigate this, we apply weight regularization, a well-established method also effective in training Constitutive Artificial Neural Networks (CANNs) \cite{mcculloch2024,linka2024}.
For finite elasticity, studies have demonstrated successful regularization strategies for CANNs.
For inelasticity, this still poses an open research question.
In contrast to elasticity, where sparsity leads to a reasonably `unique' solution, inelasticity demands not only a sparse network but also the discovery of the underlying inelastic phenomena hidden in the data. 
Ideally, if the material behaves for instance purely elastically, the network should assign zero weights to the dual potential.

We define the loss as follows
\begin{equation}
    L(\bm{S};\mathbf{W},\mathbf{b}) = \frac{1}{n_{\text{exp}}} \sum_{\alpha=1}^{n_{\text{exp}}} \frac{1}{n_{\text{data}}} \sum_{\beta=1}^{n_{\text{data}}} \| \bm{S}_\beta'(\bm{C}_\beta,\bm{U}_{i_\beta};\mathbf{W},\mathbf{b}) - \hat{\bm{S}}_\beta' \|_2^2 + R_\psi(\mathbf{W}_\psi,\mathbf{b}_\psi) + R_{\omega^*}(\mathbf{W}_{\omega^*},\mathbf{b}_{\omega^*})
\label{eq:loss}
\end{equation}
where $(\bullet)'$ refers to Voigt's notation and $\| (\bullet) \|_2^2$ refers to computing the mean of the sum of squared element-wise differences, known as L\textsubscript{2} loss.
Further, $n_{\text{exp}}$ denotes the number of experiments used for training, while $n_{\text{data}}$ is the number of data points per experiment.
The experimentally measured stress is denoted by $\hat{\bm{S}}$.
In addition, $R_\psi$ and $R_{\omega^*}$ account for the regularization of the weights and biases of the feed-forward networks of the Helmholtz free energy and the dual potential, respectively.
These weights and biases are summarized in $\mathbf{W}$ and $\mathbf{b}$.\newline

\textbf{Time integration schemes.} To solve the evolution Equation~\eqref{eq:EvolutionEquationInvars} numerically, we employ two different discretization schemes for the interval $t\in [t_{n},t_{n+1}]$\footnote{For simplicity, the subscript $n+1$ is omitted in the following.}.
The exponential integrator map is well suited in the finite strain regime as it, for instance, preserves the volume in case of an absence of hydrostatic pressure.
Moreover, as shown in \cite{holthusen2025growth}, the exponential integrator satisfies that the isochoric invariants of $\bm{U}_i$ are preserved if the deviator of $\bar{\bm{\Sigma}}$ is equal to zero.

First, we employ an explicit time integration scheme \cite{holthusen2023} 
\begin{equation}
    \bm{C}_i = \bm{U}_{i_n}\,\mathrm{exp}\left(  2\, \Delta t\, \bar{\bm{D}}_{i_n} \right)\,\bm{U}_{i_n}, \quad \bm{U}_i = + \sqrt{\bm{C}_i}
\label{eq:explicit_integration}
\end{equation}
with $\Delta t := t_{n+1}-t_n$ denoting the time increment.
Although an explicit integration is numerically efficient as no iterative solution is required, it might be less stable than an implicit scheme.
Further, we have to compute the square root of $\bm{C}_i$.
As JAX does not provide an implementation of the matrix square root at the current time, we use the generating function proposed in \cite{hudobivnik2016}.
Due to these reasons, we additionally introduce the following residual of an implicit integration scheme, cf. \cite{arghavani2011a}, \cite{holthusen2023}
\begin{equation}
    \mathrm{log}\left(\bm{U}_i^{-1}\,\bm{C}_{i_n}\,\bm{U}_i^{-1}\right) + 2\,\Delta t\, \bar{\bm{D}}_i \overset{!}{=} \bm{0}
\label{eq:implicit_integration}
\end{equation}
where $\mathrm{log}\left( \bullet \right)$ refers to the matrix logarithm, which is again computed using a generating function \cite{hudobivnik2016}.
To solve Equation~\eqref{eq:implicit_integration} in an iterative manner, we employ Broyden's method (see \ref{app:broyden}).
%
%============================================================
\subsection{Feed-forward network: Dual potential}
\label{sec:FFN_potential}
Deep feed-forward neural networks, i.e.\ multiple hidden layers between in- and outputs, serve as universal approximators \cite{hornik1989} and can approximate functions to any desired degree of accuracy if complexity of the network is increased.
Multilayer networks achieve this by recursively applying linear transformations
\begin{equation}
    \mathbf{y}_{k+1} = \mathbf{W}_{k}^{k+1}\,\mathbf{x}_k + \mathbf{b}^{k+1}
\end{equation}
from one layer, $k$, to the next layer $k+1$.
Subsequently, a nonlinear activation function, $f$, is applied on the linear transformation
\begin{equation}
    \mathbf{x}_{k+1} = f(\mathbf{y}_{k+1}).
\end{equation}
This activation function can vary between layers and can even differ between neurons in the same layer, although this is not common.
In the following, we discuss the architecture of a generalized dual potential resulting in a convex, zero-valued, and non-negative function.
Specifically, we constrain the layer stacking, permissible activation functions per layer, and the domains of weights, $\mathbf{W}$, and biases, $\mathbf{b}$, between layers.
To this end, we briefly recapture some mathematical properties of convexity, zero-valueness, and non-negativity of composed functions in Section~\ref{sec:convex_potential}.
Our discussions inspire the design of the iCANN's dual potential in Section~\ref{sec:architecture_potential}, which is capable of recovering some classical potentials from the literature, see Section~\ref{sec:resembling}.
%
%============================================================
\subsubsection{Convex, zero-valued, and non-negative dual potential}
\label{sec:convex_potential}
To ensure a thermodynamically admissible evolution of inelastic stretches, the dual potential, $\omega^*$, must be convex, non-negative, and zero-valued with respect to its arguments under any arbitrary loading.
Since we aim to discover this potential using a multilayer feed-forward neural network, we mathematically outline how such a network can be constructed. Notably, the following considerations are not `if and only if' conditions -- alternative function compositions, for example, may also yield convexity but may probably not align with the proposed framework. However, designing the network according to these principles strictly ensures thermodynamic consistency.\newline

\textbf{\textit{Convexity:} Composition of a convex function and linear combination.} Suppose we have a function $g: \mathbb{R}^n \rightarrow \mathbb{R}$ (linear combination)
\begin{equation}
	g(x_1,\hdots,x_n) = \mathbf{a}^T \mathbf{x}, \quad \mathbf{x} = \begin{pmatrix}
	x_1 \\
	\vdots \\
	x_n
\end{pmatrix},\ \mathbf{a} \in \mathbb{R}^n
\label{eq:convex_linear_1}
\end{equation}
as well as $f: \mathbb{R} \rightarrow \mathbb{R}$ which is convex.
The composition $h(\mathbf{x}) = (f \circ g)(\mathbf{x})$ is convex, since the Hessian, $H_h(\mathbf{x})$, is a rank-1 matrix of the outer product
\begin{equation}
	H_h(\mathbf{x}) = \frac{\partial^2 f}{\partial g^2}\ \mathbf{a} \mathbf{a}^T.
\label{eq:convex_linear_2}
\end{equation}
Thus, its only non-zero eigenvalue is $\frac{\partial^2 f}{\partial g^2}\mathbf{a}^T \mathbf{a}$, which is greater or equal to zero since $\frac{\partial^2 f}{\partial g^2}\geq 0$ for a convex function $f$.\newline

\textbf{\textit{Convexity:} Positive sum of convex functions.} Let us introduce a family of convex functions $f_1,f_2,\hdots,f_n:\mathbb{R}^m \rightarrow \mathbb{R}$ and $g:\mathbb{R}^m\rightarrow \mathbb{R}$ be the positive sum of these functions, i.e.,
\begin{equation}
    g(\mathbf{x}) = \sum_{i=1}^n b_i\ f_i(\mathbf{x}), \quad \mathbf{x}\in\mathbb{R}^m, \quad b_i \geq 0
\label{eq:convex_positive_sum}
\end{equation}
then $g(\mathbf{x})$ is also convex, as each $f_i$ is convex and the positive scaling of convex functions preserves their convexity. 
Noteworthy, this includes the special case where $n=1$.\newline

\textbf{\textit{Convexity:} Composition of convex functions.} We introduce two convex functions $f:\mathbb{R}\rightarrow\mathbb{R}$ and $g:\mathbb{R}^n\rightarrow\mathbb{R}$. The composition of these functions, $h = (f \circ g)(\mathbf{x})$ with $\mathbf{x}\in\mathbb{R}^n$ is convex if the outer function, $f$, is monotonically increasing.
As $g$ is itself convex, we know that $g(\theta\,\mathbf{x}_1+(1-\theta)\,\mathbf{x}_2) \leq \theta\, g(\mathbf{x}_1) + (1-\theta)\, g(\mathbf{x}_2)\ \forall\,\mathbf{x}_1,\mathbf{x}_2 \in \mathbb{R}^n$ and $\theta\in[0,1]$.
Further, since $f$ is non-decreasing, we observe the following
\begin{equation}
    f\left(g(\theta\,\mathbf{x}_1+(1-\theta)\,\mathbf{x}_2)\right) \leq f\left(\theta\, g(\mathbf{x}_1) + (1-\theta)\, g(\mathbf{x}_2)\right).
\label{eq:Composition_1}
\end{equation}
Having in mind that $f$ itself is convex, we can further conclude that
\begin{equation}
    \underbrace{f\left(\theta\, g(\mathbf{x}_1) + (1-\theta)\, g(\mathbf{x}_2)\right)}_{=(f\circ g)(\theta\,\mathbf{x}_1+(1-\theta)\,\mathbf{x}_2)} \leq \underbrace{\theta\, f\left(g(\mathbf{x}_1)\right) + (1-\theta)\, f\left(g(\mathbf{x}_2)\right)}_{\theta(f\circ g)(\mathbf{x}_1) + (1-\theta)(f\circ g)(\mathbf{x}_2)} \quad \forall\,\mathbf{x}_1,\mathbf{x}_2 \in \mathbb{R}^n,\ \theta\in[0,1]
\label{eq:Composition_2}
\end{equation}
which proves that it is sufficient to state that the composition of convex functions is convex if the outer function is non-decreasing.\newline

\textbf{\textit{Zero-valued:} Composition of zero-valued functions.} We introduce two zero-valued functions $f:\mathbb{R}^n\rightarrow\mathbb{R}$ and $g:\mathbb{R}^m\rightarrow\mathbb{R}^n$ with $f(\mathbf{0})=0$ and $g(\mathbf{0})=\mathbf{0}$. The composition of these functions is also zero-valued, i.e.
\begin{equation}
    (f \circ g)(\mathbf{0})=0.
\label{eq:zero-valued}
\end{equation}

\textbf{\textit{Non-negative:} Composition with non-negative function.} Lastly, let us define two functions $f:\mathbb{R}\rightarrow\mathbb{R}^+$ and $g:\mathbb{R}^n\rightarrow\mathbb{R}$. 
The composition of these functions is always greater than or equal to zero, i.e., 
\begin{equation}
    (f \circ g)(\mathbf{x})\geq 0  \quad \forall\,\mathbf{x} \in \mathbb{R}^n
\label{eq:non-negative}
\end{equation}
since the outer function, $f$, is always greater than or equal to zero.
\begin{remark}
As noted, a network designed according to these properties may not encompass all possible constructions of a consistent dual potential. 
However, given that neural networks are universal approximators \cite{hornik1989}, we hypothesize that increasing the number of layers and neurons enables us to discover a broad range of inelastic material behaviors.
\end{remark}
%
%============================================================
\subsubsection{Custom-designed architecture of feed-forward network}
\label{sec:architecture_potential}
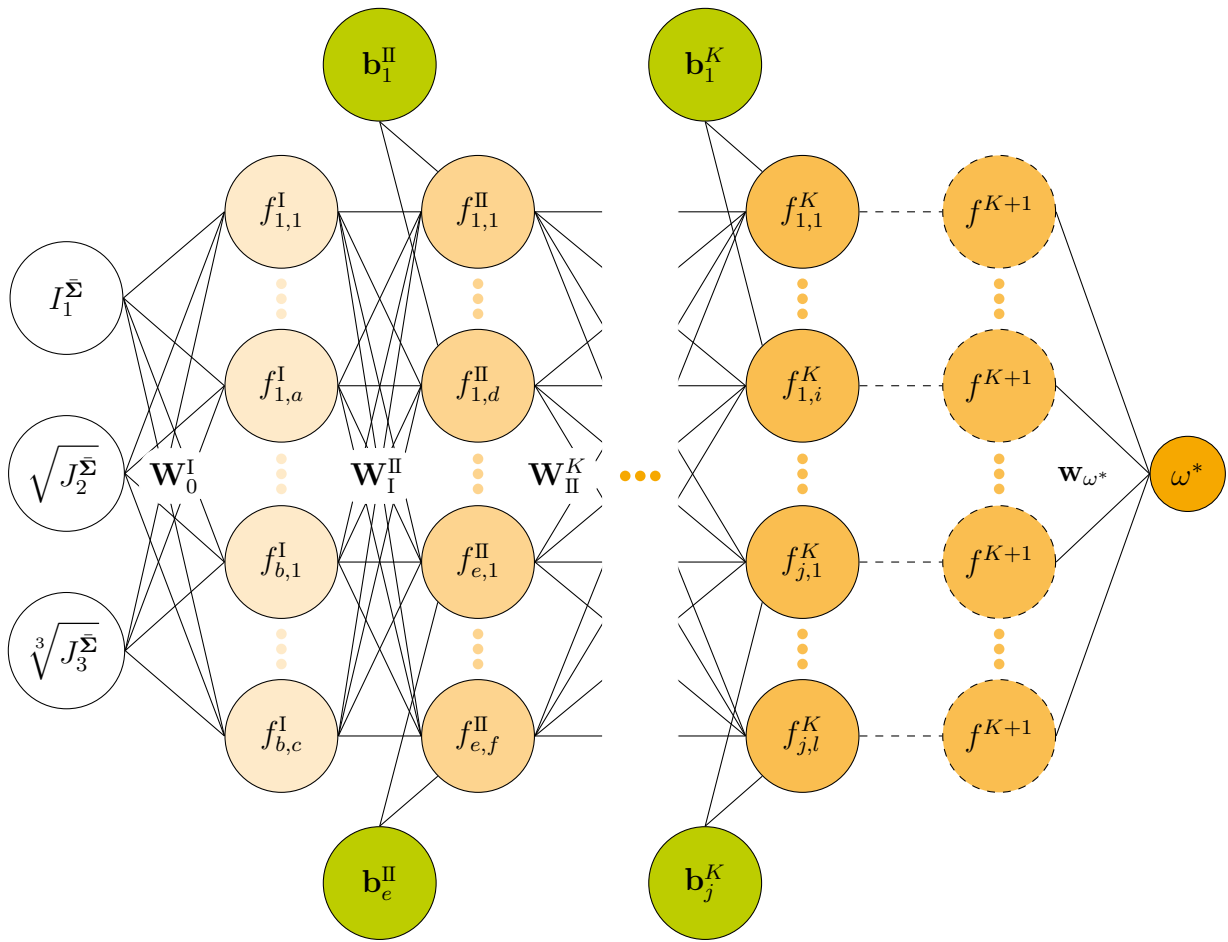
\begin{figure}[h]
    \centering
    \begin{tikzpicture}[node distance=0.8cm]
% inputs
\node (input1) at (0,0) [draw, circle, fill=white, minimum size=1.5cm] {$I_1^{\bar{\bm{\Sigma}}}$};
\node (input2) [draw, circle, fill=white, below=of input1, minimum size=1.5cm] {\hspace{-1mm}$\sqrt{J_2^{\bar{\bm{\Sigma}}}}$};
\node (input3) [draw, circle, fill=white, below=of input2, minimum size=1.5cm] {\hspace{-1mm}$\sqrt[3]{J_3^{\bar{\bm{\Sigma}}}}$};
% first hidden layer
\node (hidden2_2) [draw, circle, right = 2.1cm of $(input1)!0.5!(input2)$, minimum size = 1.5cm, fill = rwtho2 ] {$f_{1,a}^{\tightI}$};
\node (hidden2_3) [draw, circle, right = 2.1cm of $(input2)!0.5!(input3)$, minimum size = 1.5cm, fill = rwtho2 ] {$f_{b,1}^{\tightI}$};
\node (hidden2_1) [draw, circle, above =     of hidden2_2, minimum size = 1.5cm, fill = rwtho2 ] {$f_{1,1}^{\tightI}$};
\node (hidden2_4) [draw, circle, below =     of hidden2_3, minimum size = 1.5cm, fill = rwtho2 ] {$f_{b,c}^{\tightI}$};
\node (dots2_1) at ($(hidden2_1)!0.5!(hidden2_2)$) [circle, minimum size = 0.1cm ] {\textcolor{rwtho2}{\tikzvdots}};
\node (dots2_2) at ($(hidden2_2)!0.5!(hidden2_3)$) [circle, minimum size = 0.1cm ] {\textcolor{rwtho2}{\tikzvdots}};
\node (dots2_3) at ($(hidden2_3)!0.5!(hidden2_4)$) [circle, minimum size = 0.1cm ] {\textcolor{rwtho2}{\tikzvdots}};
% second hidden layer
\node (hidden3_1) [draw, circle, right = 1.1cm of hidden2_1, minimum size = 1.5cm, fill = rwtho3 ] {$f_{1,1}^{\tightII}$};
\node (hidden3_2) [draw, circle, right = 1.1cm of hidden2_2, minimum size = 1.5cm, fill = rwtho3 ] {$f_{1,d}^{\tightII}$};
\node (hidden3_3) [draw, circle, right = 1.1cm of hidden2_3, minimum size = 1.5cm, fill = rwtho3 ] {$f_{e,1}^{\tightII}$};
\node (hidden3_4) [draw, circle, right = 1.1cm of hidden2_4, minimum size = 1.5cm, fill = rwtho3 ] {$f_{e,f}^{\tightII}$};
\node (dots3_1) at ($(hidden3_1)!0.5!(hidden3_2)$) [circle, minimum size = 0.1cm ] {\textcolor{rwtho3}{\tikzvdots}};
\node (dots3_2) at ($(hidden3_2)!0.5!(hidden3_3)$) [circle, minimum size = 0.1cm ] {\textcolor{rwtho3}{\tikzvdots}};
\node (dots3_3) at ($(hidden3_3)!0.5!(hidden3_4)$) [circle, minimum size = 0.1cm ] {\textcolor{rwtho3}{\tikzvdots}};
% K-th hidden layer
\node (hiddenk_1) [draw, circle, right = 2.8cm of hidden3_1, minimum size = 1.5cm, fill = rwtho4 ] {$f_{1,1}^{K}$};
\node (hiddenk_2) [draw, circle, right = 2.8cm of hidden3_2, minimum size = 1.5cm, fill = rwtho4 ] {$f_{1,i}^{K}$};
\node (hiddenk_3) [draw, circle, right = 2.8cm of hidden3_3, minimum size = 1.5cm, fill = rwtho4 ] {$f_{j,1}^{K}$};
\node (hiddenk_4) [draw, circle, right = 2.8cm of hidden3_4, minimum size = 1.5cm, fill = rwtho4 ] {$f_{j,l}^{K}$};
\node (dotsk_1) at ($(hiddenk_1)!0.5!(hiddenk_2)$) [circle, minimum size = 0.1cm ] {\textcolor{rwtho4}{\tikzvdots}};
\node (dotsk_2) at ($(hiddenk_2)!0.5!(hiddenk_3)$) [circle, minimum size = 0.1cm ] {\textcolor{rwtho4}{\tikzvdots}};
\node (dotsk_3) at ($(hiddenk_3)!0.5!(hiddenk_4)$) [circle, minimum size = 0.1cm ] {\textcolor{rwtho4}{\tikzvdots}};
% K+1 hidden layer
\node (hiddenk1_1) [draw, dashed, circle, right = 1.1cm of hiddenk_1, minimum size = 1.5cm, fill = rwtho4 ] {$f^{K+1}$};
\node (hiddenk1_2) [draw, dashed, circle, right = 1.1cm of hiddenk_2, minimum size = 1.5cm, fill = rwtho4 ] {$f^{K+1}$};
\node (hiddenk1_3) [draw, dashed, circle, right = 1.1cm of hiddenk_3, minimum size = 1.5cm, fill = rwtho4 ] {$f^{K+1}$};
\node (hiddenk1_4) [draw, dashed, circle, right = 1.1cm of hiddenk_4, minimum size = 1.5cm, fill = rwtho4 ] {$f^{K+1}$};
\node (dotsk1_1) at ($(hiddenk1_1)!0.5!(hiddenk1_2)$) [circle, minimum size = 0.1cm ] {\textcolor{rwtho4}{\tikzvdots}};
\node (dotsk1_2) at ($(hiddenk1_2)!0.5!(hiddenk1_3)$) [circle, minimum size = 0.1cm ] {\textcolor{rwtho4}{\tikzvdots}};
\node (dotsk1_3) at ($(hiddenk1_3)!0.5!(hiddenk1_4)$) [circle, minimum size = 0.1cm ] {\textcolor{rwtho4}{\tikzvdots}};
% output
\node (output) [draw, circle, right = 2cm of $(hiddenk1_2)!0.5!(hiddenk1_3)$, minimum size = 1cm, fill = rwtho5 ] {$\omega^*$};
% bias third hidden layer
\node (bias3_1) [draw, circle, above = 1.2cm of $(hidden2_1)!0.5!(hidden3_1)$, minimum size = 1.5cm, fill = rwth6 ] {$\mathbf{b}_1^{\tightII}$};
\node (bias3_2) [draw, circle, below = 1.2cm of $(hidden2_4)!0.5!(hidden3_4)$, minimum size = 1.5cm, fill = rwth6 ] {$\mathbf{b}_e^{\tightII}$};
% bias K-th hidden layer
\node (biask_1) [draw, circle, above = 1.2cm of $(hidden3_1)!0.7!(hiddenk_1)$, minimum size = 1.5cm, fill = rwth6 ] {$\mathbf{b}_1^{K}$};
\node (biask_2) [draw, circle, below = 1.2cm of $(hidden3_4)!0.7!(hiddenk_4)$, minimum size = 1.5cm, fill = rwth6 ] {$\mathbf{b}_j^{K}$};
% connect 0 to 1
\foreach \i in {1,...,3} {
	\foreach \j in {1,...,4} {
    		\draw[-] (input\i.east) to (hidden2_\j.west);
    	}
}
% connect 1 to 2
\foreach \i in {1,...,4} {
	\foreach \j in {1,...,4} {
    		\draw[-] (hidden2_\i.east) to (hidden3_\j.west);
    	}
}
% connect 2 to K
\foreach \i in {1,...,4} {
	\foreach \j in {1,...,4} {
    		\draw[-] (hidden3_\i.east) to (hiddenk_\j.west);
    	}
}
% connect K to K+1
\foreach \i in {1,...,4} {
    \draw[-,dashed] (hiddenk_\i.east) to (hiddenk1_\i.west);
}
% connect K+1 to output
\foreach \i in {1,...,4} {
    \draw[-] (hiddenk1_\i.east) to (output.west);
}
% biases
\foreach \i in {1,...,2} {
    \draw[-] (bias3_1.south) to (hidden3_\i.north west);
    \draw[-] (biask_1.south) to (hiddenk_\i.north west);
}
\foreach \i in {3,...,4} {
    \draw[-] (bias3_2.north) to (hidden3_\i.south west);
    \draw[-] (biask_2.north) to (hiddenk_\i.south west);
}
% rectangle
\node (rec) [draw=white, fill=white, minimum width=1cm, minimum height=7.1cm, anchor=center] at ($(hidden3_2)!0.5!(hiddenk_2) + (0,-1.2)$) {};
\node (hdots) at (rec) [circle, minimum size = 0.1cm, rotate=90 ] {\textcolor{rwtho5}{\tikzvdots}}; 
% weights
\node (w01) [draw=white, fill=white, rounded corners=0.3cm] at ($(hdots)+(-6.2,0)$) {$\mathbf{W}_0^{\tightI}$};
\node (w12) [draw=white, fill=white, rounded corners=0.3cm] at ($(hdots)+(-3.5,0)$) {$\mathbf{W}_{\tightI}^{\tightII}$};
\node (w3K) [draw=white, fill=white, rounded corners=0.3cm] at ($(hdots)+(-1.1,0)$) {$\mathbf{W}_{\tightII}^{K}$};
\node (wo) [draw=white, fill=white, rounded corners=0.3cm] at ($(hdots)+(+5.9,0)$) {$\mathbf{w}_{\omega^*}$};
\end{tikzpicture}
    \caption{Schematic of our feed-forward architecture for the dual potential, $\omega^*$, designed to ensure thermodynamic consistency. The network enforces convexity, non-negativity, and a zero-valued potential. Dashed lines indicate neurons without associated weights. Activation functions are denoted as $f_{\alpha,\beta}^{\gamma}$, where $\gamma$ is the layer index, $\alpha$ specifies the function type, and $\beta$ indexes the neuron within $\alpha$. The final layer ($K+1$) exclusively employs the $\max(x,0)$ activation.
    The weights are denoted by $\mathbf{W}$ and $\mathbf{w}$, respectively, while $\mathbf{b}$ represents the network's biases.
}
    \label{fig:NN_potential_general}
\end{figure}

\textbf{General architecture.} The general architecture, whose design is deduced from Section~\ref{sec:convex_potential}, is illustrated in Figure~\ref{fig:NN_potential_general}.
We begin by refining the computation methods for the square root of $J_2^{\bar{\bm{\Sigma}}}$ and the cubic root of $J_3^{\bar{\bm{\Sigma}}}$
\begin{equation}
    \sqrt{x} := \frac{x}{(x+\epsilon_1)^{1/2}}, \quad \sqrt[3]{x} := \frac{x}{(\mathrm{abs}(x)+\epsilon_2)^{2/3}}, \quad \epsilon_1,\,\epsilon_2 \geq 0
\end{equation}
in order to be differentiable at zero.
In the numerical implementation, we choose $\epsilon_1=\epsilon_2=0.01$.

Next, let us discuss the domain of the weights and biases connecting the different layers.
According to Equations~\eqref{eq:convex_linear_1}-\eqref{eq:convex_linear_2}, the weights between the inputs and the first hidden layer are real-valued $\mathbf{W}_0^{\tightI} \in \mathbb{R}^{n_\tightI \times n_0}$. 
Here, $n_0$ refers to the number of input neurons, while $n_\tightI$ represents the total number of neurons in the first hidden layer.
For the weights between the first and second hidden layers, the matrix is given as $\mathbf{W}_{\tightI}^{\tightII} \in \mathbb{R}^{n_\tightII \times n_\tightI} \setminus S$, where
\begin{equation}
    S = \left\{ \mathbf{W}_{\tightI}^{\tightII} \in \mathbb{R}^{n_\tightII \times n_\tightI} \, : \, \exists i,j \text{ such that } W_{\tightI_{ij}}^\tightII < 0 \right\},
\end{equation}
ensuring compliance with the non-negativity constraint specified in Equation~\eqref{eq:convex_positive_sum}.
The weight matrices for subsequent layers, up to $\mathbf{W}_{\star}^{K}$, follow a similar definition. 
Further, we introduce the row vector $\mathbf{w}_{\omega^*} \in \mathbb{R}^{n_{K+1}} \setminus P$ with
\begin{equation}
    P = \left\{ \mathbf{w}_{\omega^*} \in \mathbb{R}^{n_{K+1}} \, : \, \exists i \text{ such that } w_{\omega^*_i} < 0 \right\},
\end{equation}
ensuring all weights remain non-negative.
Finally, the biases per activation function, $\mathbf{b}_\star^K$, up the second hidden layer are generally real-valued.

Following the constraints on the weights and biases of the network, we introduce how the choice of activation functions per layer and neuron is limited.
Due to Equations~\eqref{eq:convex_linear_2}-\eqref{eq:Composition_2}, we conclude that only the first hidden layer $\tightI$ permits generally convex activation functions, i.e., these functions, $f^\tightI$, might be decreasing.
Using a single activation function per layer, as is common in neural networks, severely restricts the choice of decreasing convex functions. 
To overcome this limitation, we incorporate a diverse set of activation functions in each layer, adhering to our design strategy.
The activation functions of each subsequent layer, $f^\tightII,\hdots,f^K$, are only allowed to be non-decreasing to result in a convex function, cf. Equations~\eqref{eq:convex_positive_sum} and \eqref{eq:Composition_2}.
Additionally, per Equation~\eqref{eq:zero-valued}, all activation functions must be zero-valued to achieve a zero-valued potential.

Lastly, let us consider the case where $K=\tightII$, cf. Figure~\ref{fig:NN_potential_general}.
Constructing the network with a single activation function and one neuron per layer, e.g., $f^\tightI \in \{x\}$ and $f^\tightII \in \{\mathrm{exp}(x)-1\}$, results in the potential $\omega^* = (f^\tightII \circ f^\tightI)(x) = \mathrm{exp}(w_0^\tightI \, x)-1$.
This potential is convex and zero-valued, but violates the non-negativity constraint.
To address this, we introduce an additional hidden layer, $K+1$, with the same number of neurons as layer $K$.
This final layer employs a single activation function $f^{K+1} \in \{ \mathrm{max}(x,0) \}$, cf. Equation~\eqref{eq:non-negative}.
No weights or biases are introduced between the last two hidden layers\footnote{This can be interpreted as a non-trainable identity weight matrix and a zero bias vector.}, as indicated by dashed lines in Figure~\ref{fig:NN_potential_general}.\newline

\begin{figure}[h]
    \centering
    \begin{tikzpicture}[node distance=0.8cm]
% inputs
\node (input1) at (0,0) [draw, circle, fill=white, minimum size=1.5cm] {$I_1^{\bar{\bm{\Sigma}}}$};
\node (input2) [draw, circle, fill=white, below=of input1, minimum size=1.5cm] {\hspace{-1mm}$\sqrt{J_2^{\bar{\bm{\Sigma}}}}$};
\node (input3) [draw, circle, fill=white, below=of input2, minimum size=1.5cm] {\hspace{-1mm}$\sqrt[3]{J_3^{\bar{\bm{\Sigma}}}}$};
\node (input4) [draw, circle, fill=white, below=of input3, minimum size=1.5cm] {\hspace{-1mm}$\sqrt{I_2^{\bar{\bm{\Sigma}}}}$};
\node (input5) [draw, circle, fill=white, below=of input4, minimum size=1.5cm] {\hspace{-1mm}$\sqrt[3]{I_3^{\bar{\bm{\Sigma}}}}$};
% first hidden layer
\node (hidden2_2) [draw, circle, right = 2.4cm of $(input1)!0.5!(input2)$, minimum size = 1cm, fill = rwtho2 ] {$f_{1,6}^{\tightI}$};
\node (hidden2_1) [draw, circle, above =       of hidden2_2, minimum size = 1cm, fill = rwtho2 ] {$f_{1,1}^{\tightI}$};
\node (dots2_1) at ($(hidden2_1)!0.5!(hidden2_2)$) [circle, minimum size = 0.1cm ] {\textcolor{rwtho2}{\tikzvdots}};
\node (hidden2_3) [draw, circle, right = 2.4cm of $(input2)!0.5!(input3)$, minimum size = 1cm, fill = rwtho2 ] {$f_{2,1}^{\tightI}$};
\node (hidden2_4) [draw, circle, right = 2.4cm of $(input3)!0.5!(input4)$, minimum size = 1cm, fill = rwtho2 ] {$f_{2,6}^{\tightI}$};
\node (dots2_2) at ($(hidden2_3)!0.5!(hidden2_4)$) [circle, minimum size = 0.1cm ] {\textcolor{rwtho2}{\tikzvdots}};
\node (hidden2_5) [draw, circle, right = 2.4cm of $(input4)!0.5!(input5)$, minimum size = 1cm, fill = rwtho2 ] {$f_{3,1}^{\tightI}$};
\node (hidden2_6) [draw, circle, below =       of hidden2_5, minimum size = 1cm, fill = rwtho2 ] {$f_{3,6}^{\tightI}$};
\node (dots2_3) at ($(hidden2_5)!0.5!(hidden2_6)$) [circle, minimum size = 0.1cm ] {\textcolor{rwtho2}{\tikzvdots}};
% second hidden layer
\node (hidden3_1) [draw, circle, right = 1.4cm of hidden2_2, minimum size = 1cm, fill = rwtho3 ] {$f_{1,1}^{\tightII}$};
\node (hidden3_2) [draw, circle, right = 1.4cm of hidden2_3, minimum size = 1cm, fill = rwtho3 ] {$f_{1,4}^{\tightII}$};
\node (hidden3_3) [draw, circle, right = 1.4cm of hidden2_4, minimum size = 1cm, fill = rwtho3 ] {$f_{2,1}^{\tightII}$};
\node (hidden3_4) [draw, circle, right = 1.4cm of hidden2_5, minimum size = 1cm, fill = rwtho3 ] {$f_{2,4}^{\tightII}$};
\node (dots3_1) at ($(hidden3_1)!0.5!(hidden3_2)$) [circle, minimum size = 0.1cm ] {\textcolor{rwtho3}{\tikzvdots}};
\node (dots3_2) at ($(hidden3_3)!0.5!(hidden3_4)$) [circle, minimum size = 0.1cm ] {\textcolor{rwtho3}{\tikzvdots}};
% third hidden layer
\node (hidden4_1) [draw, circle, right = 1.4cm of hidden3_1, minimum size = 1cm, fill = rwtho4 ] {$f_{1,1}^{\tightIII}$};
\node (hidden4_2) [draw, circle, right = 1.4cm of hidden3_2, minimum size = 1cm, fill = rwtho4 ] {$f_{1,4}^{\tightIII}$};
\node (hidden4_3) [draw, circle, right = 1.4cm of hidden3_3, minimum size = 1cm, fill = rwtho4 ] {$f_{2,1}^{\tightIII}$};
\node (hidden4_4) [draw, circle, right = 1.4cm of hidden3_4, minimum size = 1cm, fill = rwtho4 ] {$f_{2,4}^{\tightIII}$};
\node (dots4_1) at ($(hidden4_1)!0.5!(hidden4_2)$) [circle, minimum size = 0.1cm ] {\textcolor{rwtho4}{\tikzvdots}};
\node (dots4_2) at ($(hidden4_3)!0.5!(hidden4_4)$) [circle, minimum size = 0.1cm ] {\textcolor{rwtho4}{\tikzvdots}};
% positive hidden layer
\node (hidden5_1) [draw, dashed, circle, right = 1.4cm of hidden4_1, minimum size = 1cm, fill = rwtho4 ] {$f^{\tightIV}$};
\node (hidden5_2) [draw, dashed, circle, right = 1.4cm of hidden4_2, minimum size = 1cm, fill = rwtho4 ] {$f^{\tightIV}$};
\node (hidden5_3) [draw, dashed, circle, right = 1.4cm of hidden4_3, minimum size = 1cm, fill = rwtho4 ] {$f^{\tightIV}$};
\node (hidden5_4) [draw, dashed, circle, right = 1.4cm of hidden4_4, minimum size = 1cm, fill = rwtho4 ] {$f^{\tightIV}$};
\node (dots5_1) at ($(hidden5_1)!0.5!(hidden5_2)$) [circle, minimum size = 0.1cm ] {\textcolor{rwtho4}{\tikzvdots}};
\node (dots5_2) at ($(hidden5_3)!0.5!(hidden5_4)$) [circle, minimum size = 0.1cm ] {\textcolor{rwtho4}{\tikzvdots}};
% output
\node (output) [draw, circle, right = 2cm of $(hidden5_2)!0.5!(hidden5_3)$, minimum size = 1cm, fill = rwtho5 ] {$\omega^*$};
% bias second hidden layer
\node (bias3) [draw, circle, above = 2cm of $(hidden2_1)!0.5!(hidden3_1)$, minimum size = 1cm, fill = rwth6 ] {$\mathbf{b}_1^{\tightII}$};
% bias third hidden layer
\node (bias4) [draw, circle, right = 1cm of bias3, minimum size = 1cm, fill = rwth6 ] {$\mathbf{b}_1^{\tightIII}$};
% connect 0 to 1
\foreach \i in {1,...,5} {
	\foreach \j in {1,...,6} {
    		\draw[-] (input\i.east) to (hidden2_\j.west);
    	}
}
% connect 1 to 2
\foreach \i in {1,...,6} {
	\foreach \j in {1,...,4} {
    		\draw[-] (hidden2_\i.east) to (hidden3_\j.west);
    	}
}
% connect 2 to 3
\foreach \i in {1,...,4} {
	\foreach \j in {1,...,4} {
    		\draw[-] (hidden3_\i.east) to (hidden4_\j.west);
    	}
}
% connect 3 to 4
\foreach \i in {1,...,4} {
    \draw[-,dashed] (hidden4_\i.east) to (hidden5_\i.west);
}
% connect 4 to output
\foreach \i in {1,...,4} {
    \draw[-] (hidden5_\i.east) to (output.west);
}
% biases
\foreach \i in {1,...,2} {
    \draw[-] (bias3.south) to (hidden3_\i.north west);
    \draw[-] (bias4.south) to (hidden4_\i.north west);
}
% weights
\node (w01) [draw=white, fill=white, rounded corners=0.3cm] at ($(input3)+(+1.6,0)$) {$\mathbf{W}_0^{\tightI}$};
\node (w12) [draw=white, fill=white, rounded corners=0.3cm] at ($(input3)+(+4.3,0)$) {$\mathbf{W}_{\tightI}^{\tightII}$};
\node (w23) [draw=white, fill=white, rounded corners=0.3cm] at ($(input3)+(+6.9,0)$) {$\mathbf{W}_{\tightII}^{\tightIII}$};
\node (wo) [draw=white, fill=white, rounded corners=0.3cm] at ($(input3)+(+11.8,0)$) {$\mathbf{w}_{\omega^*}$};
\end{tikzpicture}
    \caption{Specification of our general architecture presented in Figure~\ref{fig:NN_potential_general}. The inputs are extended by two invariants, while biases are only associated with the maximum activation function in the second and third layer.}
    \label{fig:NN_potential}
\end{figure}
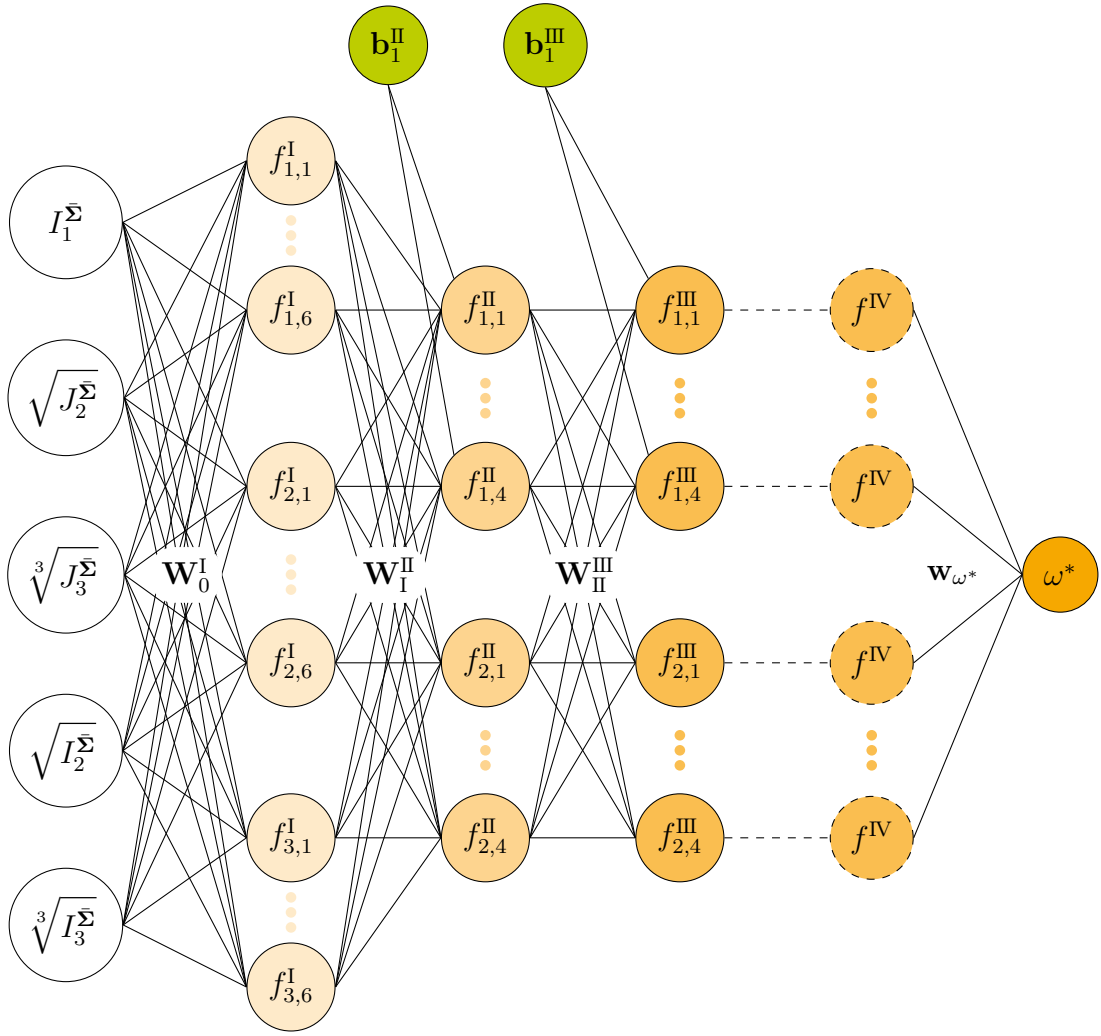

\textbf{Specific architecture.} With the general architecture at hand, we need to specify the number of layers, the choice of activation functions per layer, and the number of neurons per activation function.
In this contribution, we employ the specific architecture shown in Figure~\ref{fig:NN_potential}.
First of all, we enhance the network's inputs by the stress invariants $I_2^{\bar{\bm{\Sigma}}} := 1/2\, \mathrm{tr}(\bar{\bm{\Sigma}}^2)$ and $I_3^{\bar{\bm{\Sigma}}} := 1/3\, \mathrm{tr}(\bar{\bm{\Sigma}}^3)$.
This is an easy method of broadening the space of potentials included in our network, see also Section~\ref{sec:resembling}.
Note that this does not affect our findings on the potential, see \ref{app:stress_invars}.

In line with the design of the architecture introduced above, we choose the following set of activation functions per hidden layer
\begin{equation}
    \begin{aligned}
        f^{\tightI} &\in \{x,|x|^{p_1+1},\mathrm{ln}(\mathrm{cosh}(|x|^{p_2+1}))\} \\
        f^{\tightII} &\in \{\mathrm{max}(x,0),\mathrm{exp}(x)-1\} \\
        f^{\tightIII} &\in \{\mathrm{max}(x,0),\mathrm{exp}(x)-1\}.
    \end{aligned}
\end{equation}
Here, we introduce two additional weights, $p_1$ and $p_2$, which are discovered during training.
These weights are constrained to be greater than or equal to zero, i.e.\ $p_1,\, p_2 \geq 0$, to obtain convex activation functions.
Moreover, we modify the computation of $|x|^{\alpha+1}$ to guarantee that the function is finite for $x=0$ and differentiable at $x=0$
\begin{equation}
    |x|^{\alpha+1} := \mathrm{exp}\left(\left(\alpha + 1\right) \ln\left(\mathrm{abs}(u(x)) + \epsilon_1\right)\right),\quad u(x) = x + \mathrm{sgn}(x)\,\epsilon_2, \quad \epsilon_1,\,\epsilon_2 \geq 0
\end{equation}
where $\mathrm{sgn}(x)$ refers to the sign function.
In the numerical implementation, we choose $\epsilon_1 = \epsilon_2 = 0.0001$.
In addition, we introduce biases, $\mathbf{b}_1^{\tightII}$ and $\mathbf{b}_1^{\tightIII}$, for the $\mathrm{max}(x,0)$ activation functions in the second and third layers, respectively.
Noteworthy, the entries of theses biases must be non-positive to satisfy that the activation function remains zero-valued.

It remains to choose the number of neurons per activation function.
For simplicity, we choose the same number of neurons per activation function in each layer.
In the first layer, we employ six neurons per function resulting in eighteen neurons in total.
For all subsequent layers, we choose four neurons per activation function.
Consequently, our specific architecture consist of $90\,(\mathbf{W}_0^\tightI) + 144\,(\mathbf{W}_\tightI^\tightII) + 64\,(\mathbf{W}_\tightII^\tightIII) + 8\,(\mathbf{w}_{\omega^*}) + 4\,(\mathbf{b}_1^\tightII) + 4\,(\mathbf{b}_1^\tightIII) + 1\, (p_1) + 1\, (p_2) = 316$ weights in total.\newline

\textbf{Regularization.} In Section~\ref{sec:network}, we introduced the training loss with the regularization $R_{\omega^*}$ for the potential.
Furthermore, we stated that the regularization should discover a relatively sparse network in order to both be considered as a `unique' material model and to reveal the inelastic phenomena.
Thus, in line with the experiences made in \cite{mcculloch2024,linka2024}, we regularize the weights of the last layer, $\mathbf{w}_{\omega^*}$, by a lasso (L\textsubscript{1}) regularization to promote sparsity.
Additionally, we regularize the weights of the very first hidden layer by an elastic net, which linearly combines lasso and ridge (L\textsubscript{2}) regularization, to promote sparsity and mitigate multicollinearity.
Lastly, the biases are regularized by lasso.
Thus, the overall regularization of the potential reads
\begin{equation}
    R_{\omega^*} = \sum_{i=1}^{18}\sum_{j=1}^{5}\left[\lambda_1\,\left(W_{0_{ij}}^\tightI\right)^2 + \lambda_2\,\mathrm{abs}\left(W_{0_{ij}}^\tightI\right)\right] 
    +\sum_{i=1}^{8} \lambda_3\,\mathrm{abs}\left(w_{\omega^*_{i}}\right) +\sum_{i=1}^{4} \lambda_4\,\left[\mathrm{abs}\left(b_{1_i}^\tightII\right) + \mathrm{abs}\left(b_{1_i}^\tightIII\right) \right]
\label{eq:regularization_potential}
\end{equation}
where we set the regularization parameters to $\lambda_1=\lambda_2=\lambda_3=0.0001$ and $\lambda_4=0.01$ for all training sessions.

%============================================================
\subsubsection{Reproduction of classical dual potentials}
\label{sec:resembling}
\begin{figure}[h]
    \centering
    \begin{tikzpicture}[node distance=0.4cm]
% inputs
\node (input1) at (0,0) [draw, circle, fill=white, minimum size=1.5cm] {$I_1^{\bar{\bm{\Sigma}}}$};
\node (input2) [draw, circle, fill=white, below=of input1, minimum size=1.5cm] {\hspace{-1mm}$\sqrt{J_2^{\bar{\bm{\Sigma}}}}$};
\node (input3) [draw, circle, fill=white, below=of input2, minimum size=1.5cm] {\hspace{-1mm}$\sqrt{I_2^{\bar{\bm{\Sigma}}}}$};
% first hidden layer
\node (hidden1_1) [draw, circle, right = 2.0cm of input1, minimum size = 1cm, fill = rwtho2 ] {$f_{1,1}^{\tightI}$};
\node (hidden1_2) [draw, circle, right = 2.0cm of input2, minimum size = 1cm, fill = rwtho2 ] {$f_{2,1}^{\tightI}$};
\node (hidden1_3) [draw, circle, right = 2.0cm of input3, minimum size = 1cm, fill = rwtho2 ] {$f_{2,2}^{\tightI}$};
% second hidden layer
\node (hidden3_1) [draw, circle, right = 1.2cm of hidden1_2, minimum size = 1cm, fill = rwtho3 ] {$f_{1,1}^{\tightII}$};
% third hidden layer
\node (hidden4_1) [draw, circle, right = 1.6cm of hidden3_1, minimum size = 1cm, fill = rwtho4 ] {$f_{1,1}^{\tightIII}$};
% positive hidden layer
\node (hidden5_1) [draw, dashed, circle, right = 1.6cm of hidden4_1, minimum size = 1cm, fill = rwtho4 ] {$f^{\tightIV}$};
% output
\node (output) [draw, circle, right = 1.6cm of hidden5_1, minimum size = 1cm, fill = rwtho5 ] {$\omega^*$};
% connect 0 to 1
\foreach \i in {1,...,3} {
	\foreach \j in {1,...,3} {
    		\draw[-] (input\i.east) to (hidden1_\j.west);
    	}
}
% connect 1 to 2
\foreach \i in {1,...,3} {
    \draw[-] (hidden1_\i.east) to (hidden3_1.west);
}
% connect 2 to 3
\draw[-] (hidden3_1.east) to node[above] {$w_{\tightII}^{\tightIII}=1$} (hidden4_1.west);
% connect 3 to 4
\draw[-,dashed] (hidden4_1.east) to (hidden5_1.west);
% connect 3 to output
\draw[-] (hidden5_1.east) to node[above] {$w_{\omega^*}=1$} (output.west);
% weights
\node (w01) [draw=white, fill=white, rounded corners=0.3cm] at ($(input2)+(+1.8,0)$) {$\mathbf{W}_0^{\tightI}$};
\node (w12) [draw=white, fill=white, rounded corners=0.3cm] at ($(input2)+(+4.6,0)$) {$\mathbf{w}_{\tightI}^{\tightII}$};
\end{tikzpicture}
    \caption{Reduced architecture of the specific network shown in Figure~\ref{fig:NN_potential}. For simplicity, the weights between the second and third layers as well as the third layer to the output are set equal to one. The activation functions are chosen as $f^\tightI \in \{x,|x|^{p_1+1}\}$ and $f^\tightII,\, f^\tightIII \in \{\mathrm{max}(x,0)\}$, where the weight $p_1$ is set equal to one.}
    \label{fig:NN_resembling}
\end{figure}
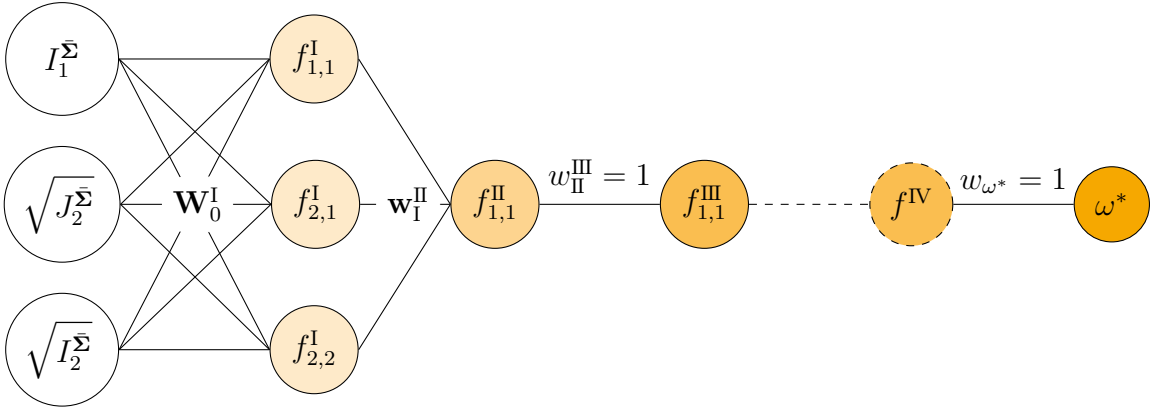

In the following, we demonstrate that the generalized iCANN architecture, as previously introduced, is capable of recovering a variety of established dual potentials known within the continuum mechanics community. 
To this end, we streamline the architecture outlined in Section~\ref{sec:architecture_potential} to its essential neurons, thereby enabling the integration of all these models into a unified architecture. 
The resulting reduced network is illustrated in Figure~\ref{fig:NN_resembling}. 
For each of the classical dual potentials examined, we link weight matrices to the material parameters.

The \textit{von Mises} criterion \cite{vonmises1913} can be expressed as
\begin{equation}
    \omega^*_{vM} = \sqrt{3\, J_2^{\bar{\bm{\Sigma}}} }, \quad \mathbf{W}_0^\tightI = \begin{pmatrix}
        0 & 1 & 0 \\
        0 & 0 & 0 \\
        0 & 0 & 0
    \end{pmatrix}, \quad \mathbf{w}_\tightI^\tightII = \begin{pmatrix}
        \sqrt{3} & 0 & 0
    \end{pmatrix}
\end{equation}
which is mostly applied for ductile materials.
At this point, it is already important to note that due to the dense structure of the generalized architecture, the weight matrices are not unique.
For instance, for the \textit{von Mises} potential one may exchange the non-zero entries in $\mathbf{W}_0^\tightI$ with those of $\mathbf{w}_\tightI^\tightII$.

Next, we investigate the pressure-sensitive \textit{Drucker-Prager} model \cite{drucker1952}
\begin{equation}
    \omega^*_{DP} = \sqrt{3\, J_2^{\bar{\bm{\Sigma}}} } + \underbrace{\frac{\sigma_c-\sigma_t}{\sigma_c+\sigma_t}}_{=:\xi}\, I_1^{\bar{\bm{\Sigma}}}, \quad \mathbf{W}_0^\tightI = \begin{pmatrix}
        \xi & \sqrt{3} & 0 \\
        0 & 0 & 0 \\
        0 & 0 & 0
    \end{pmatrix}, \quad \mathbf{w}_\tightI^\tightII = \begin{pmatrix}
        1 & 0 & 0
    \end{pmatrix}
\end{equation}
which represents a cone within the principal stress space.
The yield stresses, $\sigma_c$ and $\sigma_t$, denote the uniaxial yield stress in compression and tension, respectively.
It is important to note that $\xi$ might be negative.
Note that using the `original' iCANN architecture \cite{holthusen2024}, we were not able to discover the \textit{Drucker-Prager} model, cf. \cite{boes2024arxiv}.

The \textit{Bresler-Pister} \cite{bresler1985} model extends the former potential by a dependence on the quadratic hydrostatic pressure $I_1^{\bar{\bm{\Sigma}}}$.
Usually, this additional dependency is associated with the yield stress under biaxial compression.
The potential reads as follows
\begin{equation}
    \omega^*_{BP} = \sqrt{3\, J_2^{\bar{\bm{\Sigma}}} } + \zeta_1\,I_1^{\bar{\bm{\Sigma}}} + \zeta_2\,\left( I_1^{\bar{\bm{\Sigma}}} \right)^2, \quad \mathbf{W}_0^\tightI = \begin{pmatrix}
        \zeta_1 & \sqrt{3} & 0 \\
        1 & 0 & 0 \\
        0 & 0 & 0
    \end{pmatrix}, \quad \mathbf{w}_\tightI^\tightII = \begin{pmatrix}
        1 & \zeta_2 & 0
    \end{pmatrix}
\end{equation}
with the material constants $\zeta_1$ and $\zeta_2$.
As for the \textit{Drucker-Prager} model, $\zeta_1$ can be any real number.
In contrast, $\zeta_2$ must be positive to be discoverable by the generalized iCANN.

The potential proposed by \cite{stassi1967,tschoegl1971} represents a paraboloid within the principal stress space
\begin{equation}
    \omega^*_{ST} = 3\, J_2^{\bar{\bm{\Sigma}}} + (\sigma_c-\sigma_t)\, I_1^{\bar{\bm{\Sigma}}}, \quad \mathbf{W}_0^\tightI = \begin{pmatrix}
        \sigma_c-\sigma_t & 0 & 0 \\
        0 & \sqrt{3} & 0 \\
        0 & 0 & 0
    \end{pmatrix}, \quad \mathbf{w}_\tightI^\tightII = \begin{pmatrix}
        1 & 1 & 0
    \end{pmatrix}
\end{equation}
where the difference, $\sigma_c-\sigma_t$, can be negative.
Again, this potential cannot be discovered by the original iCANN architecture, see \cite{boes2024arxiv}.

A quadratic potential, which is often used to model visco-elasticity at finite strains \cite{reese1998}, is expressed in terms of stress invariants as follows
\begin{equation}
    \omega^*_{RG} = 0.25\,\mu^{-1}\, J_2^{\bar{\bm{\Sigma}}} + \frac{\kappa^{-1}}{18}\, \left( I_1^{\bar{\bm{\Sigma}}} \right)^2, \quad \mathbf{W}_0^\tightI = \begin{pmatrix}
        0 & 0 & 0 \\
        1 & 0 & 0 \\
        0 & 1 & 0
    \end{pmatrix}, \quad \mathbf{w}_\tightI^\tightII = \begin{pmatrix}
        0 & \kappa^{-1}/18 & 0.25\,\mu^{-1} 
    \end{pmatrix}
\end{equation}
where $\mu$ denotes the material's shear modulus, while $\kappa$ refers to the bulk modulus.

Lastly, we investigate a smoothed version \cite{holthusen2023} of the maximum principal stress theory, also known as Rankine's theory \cite{collins1993book},
\begin{equation}
    \omega^*_{sPs} = I_1^{\bar{\bm{\Sigma}}} + \sqrt{ I_2^{\bar{\bm{\Sigma}}} }, \quad \mathbf{W}_0^\tightI = \begin{pmatrix}
        1 & 0 & 1 \\
        0 & 0 & 0 \\
        0 & 0 & 0
    \end{pmatrix}, \quad \mathbf{w}_\tightI^\tightII = \begin{pmatrix}
        1 & 0 & 0
    \end{pmatrix}
\end{equation}
where it becomes clear why we include also the invariant $I_2^{\bar{\bm{\Sigma}}}$ into the specific architecture of our proposed generalized dual potential.
%
%============================================================
\subsection{Feed-forward network: Helmholtz free energy}
\label{sec:FFN_energy}
\begin{figure}[h]
    \centering
    \begin{tikzpicture}[node distance=1.3cm]
% inputs
\node (input1) at (0,0) [draw, circle, fill=white, minimum size=2cm] {$\tilde{I}_1^{\bar{\bm{C}}_e}$};
\node (input2) [draw, circle, fill=white, below= of input1, minimum size=2cm] {$\left(\tilde{I}_2^{\bar{\bm{C}}_e}\right)^{\frac{3}{2}}$};
% first hidden layer
\node (hidden1_2) [draw, circle, right = 2cm of input1, minimum size = 1cm, fill = rwthb2 ] {$f_{1,2}^{\tightI}$};
\node (hidden1_1) [draw, circle, above =     of hidden1_2, minimum size = 1cm, fill = rwthb2 ] {$f_{1,1}^{\tightI}$};
\node (hidden1_3) [draw, circle, right = 2cm of input2, minimum size = 1cm, fill = rwthb2 ] {$f_{2,1}^{\tightI}$};
\node (hidden1_4) [draw, circle, below =     of hidden1_3, minimum size = 1cm, fill = rwthb2 ] {$f_{2,2}^{\tightI}$};
\node (hidden1_5) [draw, circle, below = 1cm of hidden1_4, minimum size = 1cm, fill = rwthb2 ] {$f_{2,3}^{\tightI}$};
% second hidden layer
\node (hidden2_8) [draw, circle, right = 2cm of hidden1_4, minimum size = 1cm, fill = rwthb3 ] {$f_{2,4}^{\tightII}$};
\node (hidden2_6) [draw, circle, right = 2cm of hidden1_3, minimum size = 1cm, fill = rwthb3 ] {$f_{2,2}^{\tightII}$};
\node (hidden2_7) at ($(hidden2_8)!0.5!(hidden2_6)$) [draw, circle, minimum size = 1cm, fill = rwthb3 ] {$f_{2,3}^{\tightII}$};
\node (hidden2_5) [draw, circle, right = 2cm of hidden1_2, minimum size = 1cm, fill = rwthb3 ] {$f_{1,4}^{\tightII}$};
\node (hidden2_3) [draw, circle, right = 2cm of hidden1_1, minimum size = 1cm, fill = rwthb3 ] {$f_{1,2}^{\tightII}$};
\node (hidden2_4) at ($(hidden2_5)!0.5!(hidden2_3)$) [draw, circle, minimum size = 1cm, fill = rwthb3 ] {$f_{1,3}^{\tightII}$};
\node (hidden2_2) at ($(hidden2_5)!0.5!(hidden2_6)$) [draw, circle, minimum size = 1cm, fill = rwthb3 ] {$f_{2,1}^{\tightII}$};
\node (hidden2_1) [draw, circle, above = 0.05cm of hidden2_3, minimum size = 1cm, fill = rwthb3 ] {$f_{1,1}^{\tightII}$};
\node (hidden2_9) [draw, circle, right = 2cm of hidden1_5, minimum size = 1cm, fill = rwthb3 ] {$f_{3,1}^{\tightII}$};
% inputs
\node (input3) [draw, circle, fill=white, left= 2cm of hidden1_5, minimum size=2cm] {$I_3^{\bar{\bm{C}}_e}$};
% output
\node (output) [draw, circle, right = 2cm of hidden2_6, minimum size = 1cm, fill = rwthb4 ] {$\psi$};
% bias second hidden layer
\node (bias1) [draw, circle, above = 1cm of $(input1)!0.5!(hidden1_1)$, minimum size = 1.2cm, fill = rwth6 ] {$-3$};
\node (bias2) [draw, circle, below = 1cm of $(input2)!0.5!(hidden1_4)$, minimum size = 1.2cm, fill = rwth6 ] {$-3^{\frac{3}{2}}$};
% connections
% 0 to 1
\draw[-,dashed] (input1.east) to (hidden1_1.west);
\draw[-,dashed] (input1.east) to (hidden1_3.west);
\draw[-,dashed] (input2.east) to (hidden1_2.west);
\draw[-,dashed] (input2.east) to (hidden1_4.west);
\draw[-,dashed] (input3.east) to (hidden1_5.west);
% 1 to 2
\draw[-,dashed] (hidden1_1.east) to (hidden2_1.west);
\draw[-] (hidden1_1.east) to (hidden2_2.west);
\draw[-,dashed] (hidden1_2.east) to (hidden2_3.west);
\draw[-] (hidden1_2.east) to (hidden2_6.west);
\draw[-,dashed] (hidden1_3.east) to (hidden2_4.west);
\draw[-] (hidden1_3.east) to (hidden2_7.west);
\draw[-,dashed] (hidden1_4.east) to (hidden2_5.west);
\draw[-] (hidden1_4.east) to (hidden2_8.west);
\draw[-,dashed] (hidden1_5.east) to (hidden2_9.west);
% bias
\draw[-,dashed] (bias1.east) to (hidden1_1.north west);
\draw[-,dashed] (bias1.east) to (hidden1_3.north west);
\draw[-,dashed] (bias2.east) to (hidden1_2.south west);
\draw[-,dashed] (bias2.east) to (hidden1_4.south west);
% 2 to output
\foreach \i in {1,...,9} {
    \draw[-] (hidden2_\i.east) to (output.west);
}
% weights
\node (wo) [draw=white, fill=white, rounded corners=0.3cm] at ($(output)+(-1.5,0)$){$\mathbf{w}_{\psi}$};
\node (w12) [draw=white, fill=white, rounded corners=0.3cm] at ($(output)+(-4.7,0)$){$\mathbf{W}_\tightI^\tightII$};
\end{tikzpicture}
    \caption{Schematic of our feed-forward architecture for the Helmholtz free energy, $\psi$. Dashed lines indicate neurons without associated weights, respectively, being equal to one and non-trainable. The biases are constant and are introduced to ensure a zero-valued energy.}
    \label{fig:NN_energy}
\end{figure}
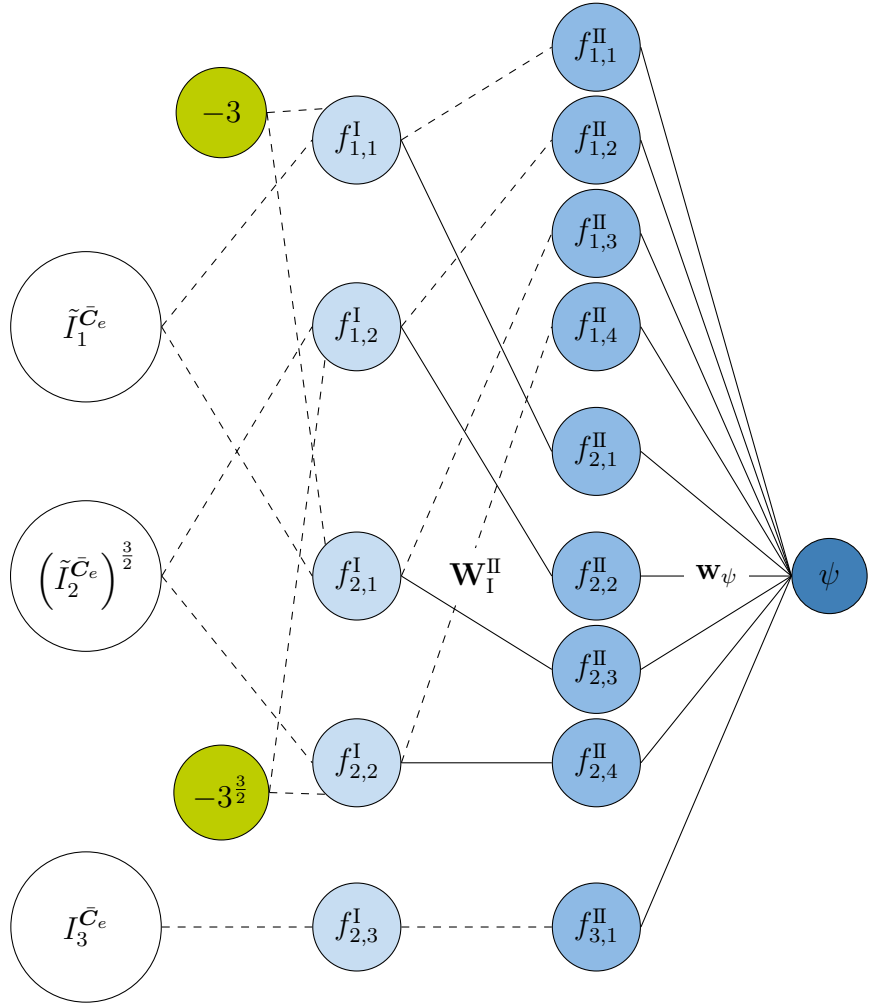

\textbf{Architecture.} As the scope of this manuscript lies on the development of a generalized dual potential, we employ the feed-forward network of the Helmholtz free energy proposed in \cite{holthusen2024,holthusen2024PAMM}.
Figure~\ref{fig:NN_energy} illustrates the custom-designed feed-forward network.
In contrast to our new proposed network of the potential, the energy network is sparse.
As $\psi$ is an isotropic function of $\bar{\bm{C}}_e$, we express it in terms of invariants.
To this end, the determinant, $I_3^{\bar{\bm{C}}_e}:=\mathrm{det}(\bar{\bm{C}}_e)$, as well as the isochoric invariants, $\tilde{I}_1^{\bar{\bm{C}}_e} := \mathrm{tr}(\bar{\bm{C}}_e)/(I_3^{\bar{\bm{C}}_e})^{1/3}$ and $\tilde{I}_2^{\bar{\bm{C}}_e} := \mathrm{tr}(\mathrm{cof}(\bar{\bm{C}}_e))/(I_3^{\bar{\bm{C}}_e})^{2/3}$, serve as the network's inputs.
The latter invariant is raised to the power of $3/2$ to satisfy polyconvexity \cite{hartmann2003}.

For the network, we choose the following set of activation functions per layer
\begin{equation}
    \begin{aligned}
        f^{\tightI} &\in \{x^2,x\} \\
        f^{\tightII} &\in \{x,\mathrm{exp}(x)-1,x^{p_3}-1-\mathrm{ln}(x^{p_3})\}
    \end{aligned}
\end{equation}
where we introduce the weight $p_3$ in addition to the weights of the network, see Figure~\ref{fig:NN_energy}.
Thus, the total number of weights is $4\, (\mathbf{W}_\tightI^\tightII) + 9\, (\mathbf{w}_\psi) + 1\, (p_3) = 14$ for the network of the energy.\newline

\textbf{Regularization.} As for the dual potential, we regularize the weights introduced for the energy (cf. Equations~\eqref{eq:loss} and \eqref{eq:regularization_potential}).
According to the experiences made in \cite{holthusen2023PAMM,linka2024,holthusen2024}, we only employ lasso regularization for the weights of the very last layer, and only regularize the weights associated with the isochoric part of the energy, i.e.,
\begin{equation}
    R_\psi = \sum_{i=1}^8 \lambda_5\, \mathrm{abs}\left( w_{\psi_i} \right)
\end{equation}
where $\lambda_5$ denotes the regularization parameter which is set to $0.001$ within this contribution.

\section{Results}
\label{sec:results}
In the following, we systematically train our generalized iCANN model and evaluate its predictive accuracy on both synthetic (Section~\ref{sec:results_artificial}) and experimentally acquired data (Section~\ref{sec:Hossain2012}). Additionally, we compare explicit and implicit integration schemes, implement a staggered discovery approach, and introduce noise into the data set to assess potential overfitting effects.

To this end, we first consider two distinct artificially generated data sets in Sections~\ref{sec:results_artificial_1} and \ref{sec:results_artificial_2}. These data sets differ in terms of the underlying dual potential as well as the magnitude of dissipated energy. To validate our generalized iCANN model against experimental data, we infer the visco-elastic response of VHB 4910 polymer \cite{hossain2012}.

\begin{figure}[h]
    \centering
    \begin{tikzpicture}
    % Spring 1
    \draw[-] (0,0) -- (1,0) -- (1,0.5) -- (1.5,-0.5) -- (1.5,0.5) -- (2,-0.5) -- (2,0.5) -- (2.25,0) -- (3,0);
    % dashpot 1
    \draw[-] (3,0) -- (3.5,0) -- (3.5,0.5) -- (3.5,-0.5);
    \draw[-] (3.25,0.5) -- (4,0.5) -- (4,-0.5) -- (3.25,-0.5);
    \draw[-] (4,0) -- (5,0);
    % Spring 2
    \draw[-] (0,0-2) -- (1,0-2) -- (1,0.5-2) -- (1.5,-0.5-2) -- (1.5,0.5-2) -- (2,-0.5-2) -- (2,0.5-2) -- (2.25,0-2) -- (3,0-2);
    % dashpot 2
    \draw[-] (3,0-2) -- (3.5,0-2) -- (3.5,0.5-2) -- (3.5,-0.5-2);
    \draw[-] (3.25,0.5-2) -- (4,0.5-2) -- (4,-0.5-2) -- (3.25,-0.5-2);
    \draw[-] (4,0-2) -- (5,0-2);
    % left boundary
    \draw[-,line width = 3pt] (0,0.5) -- (0,-2.5);
    % right boundary
    \draw[-] (5,0) -- (5,0-2);
    \draw[->,line width = 3pt] (5,-1) -- (6.0,-1);
    % description
    \node (psi1) [] at (1.5,1.0) {$\prescript{1}{}{}\psi\left(\prescript{1}{}{}\bar{\bm{C}}_e\right)$};
    \node (psi2) [] at (1.5,-3.0) {$\prescript{2}{}{}\psi\left(\prescript{2}{}{}\bar{\bm{C}}_e\right)$};
    \node (w1) [] at (3.75,1.0) {$\prescript{1}{}{}\omega^*\left(\prescript{1}{}{}\bar{\bm{\Sigma}}\right)$};
    \node (w2) [] at (3.75,-3.0) {$\prescript{2}{}{}\omega^*\left(\prescript{2}{}{}\bar{\bm{\Sigma}}\right)$};
    \node (S) [anchor=west] at (6.2,-1) {$\bm{S} = \prescript{1}{}{}\bm{S} + \prescript{2}{}{}\bm{S}$};
\end{tikzpicture}
    \caption{Rheological representation of the parallel connection of two generalized iCANNs. Each iCANN consists of an individual `spring', $\prescript{1}{}{}\psi$ or $\prescript{2}{}{}\psi$, and `dashpot', $\prescript{1}{}{}\omega^*$ or $\prescript{2}{}{}\omega^*$, element with individual states, $\prescript{1}{}{}\bm{U}_i$ and $\prescript{2}{}{}\bm{U}_i$, respectively. The entire Helmholtz free energy is $\psi = \prescript{1}{}{}\psi + \prescript{2}{}{}\psi$, while the overall dual potential is the sum of the individual potentials, i.e., $\omega^*=\prescript{1}{}{}\omega^* + \prescript{2}{}{}\omega^*$. For both individual iCANNs, the same right Cauchy-Green tensor, $\bm{C}$, serves as input. Consequently, the outputs are summed up to a single second Piola-Kirchhoff stress, $\bm{S}$.}
    \label{fig:rheo}
\end{figure}
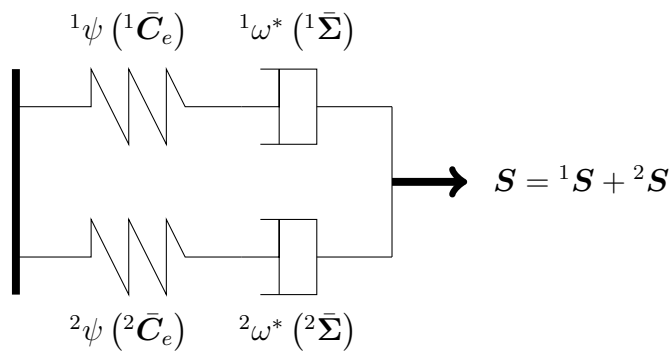

Figure~\ref{fig:rheo} schematically represents the iCANN architecture in terms of a rheological model. 
Since a standard Maxwell element, consisting of a series connection of an `elastic spring' and a `viscous dashpot', is insufficient to characterize visco-elastic solids, we employ a parallel configuration of two Maxwell elements ($\prescript{1}{}{}(\bullet)$ and $ \prescript{2}{}{}(\bullet)$). 
This configuration is intended to facilitate the identification of the underlying visco-elastic behavior through training.

For all training sessions, we initialize the network's weights and biases using a uniform random distribution. 
We maintain fixed lower and upper value bounds across all training scenarios, rather than tuning them to specific problem instances. 
Furthermore, our regularization parameters remain constant throughout training. 
The mean squared error serves as the loss function, quantifying the discrepancy between the training data and the neural network’s predictions.

The entire computational framework is implemented in JAX \cite{jax2018github}, utilizing the ADAM optimizer \cite{kingma2017adammethodstochasticoptimization} from the Optax library with a learning rate of $0.001$. 
Additionally, we apply gradient clipping based on a maximum global norm, set to $0.001$.
Our training experience made during the preparation of this study indicates that gradient clipping is particularly crucial for inelastic materials, as it prevents excessive growth in the potential’s weights. 
Uncontrolled weight growth can lead to an unstable evolution equation, potentially resulting in unbounded stress values.
%
%================================================
\subsection{Discovery of artificial data}
\label{sec:results_artificial}
\begin{figure}[h]
    \centering
    \includegraphics[]{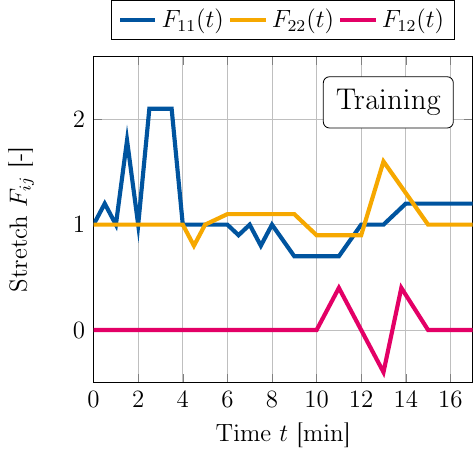}
    \includegraphics[]{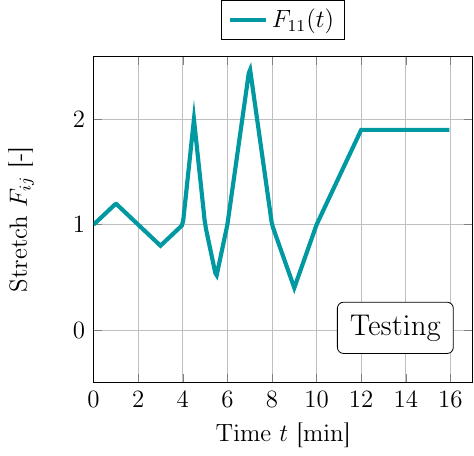}
    \caption{Left: Deformation gradient components used for creating a training data set. The third entry on the main diagonal, $F_{33}$, is set to one, while the remaining shear terms are zero. The training set includes multiaxial, cyclic, and relaxation behavior. Right: Prescribed stretch in the main stretch direction for uniaxial loading employed for testing. The remaining components of the deformation gradient are computed according to the constraint that the entries of $\bm{S}$ in the off-main diagonal direction are zero.}
    \label{fig:loading}
\end{figure}

In line with Shannon's information theory \cite{shannon1948}, `rich' data sets should enable us to discover a `unique' set of weights and biases; however, `rich' does not necessarily mean more data points.
Key features such as diverse data sets and higher information entropy prevent overfitting, while the performance during gradient optimization usually improves.
In terms of material science and computational discovery, we hypothesize that `rich' data sets featuring complex, relaxing, multiaxial, and cyclic load paths will help us to discover our weights in the network.
Thus, we prescribe the loading path shown in Figure~\ref{fig:loading} with $\Delta t = 0.05$ [min] to create a `rich' data set, which is utilized to obtain stress-time data in Sections~\ref{sec:results_artificial_1} and \ref{sec:results_artificial_2}.
To test our discovered weights and biases, we subject the network to uniaxial loading.
Here, we compute the remaining entries in Figure~\ref{fig:loading} (loading) according to the uniaxial stress constraint with $\Delta t = 0.06$ [min].
Noteworthy, the components of the deformation gradient may vary depending on the energy and potential employed.
%
%================================================
\subsubsection{Discovering a model for artificial data: Example 1}
\label{sec:results_artificial_1}
\begin{figure}[h]
    \centering
    \includegraphics[]{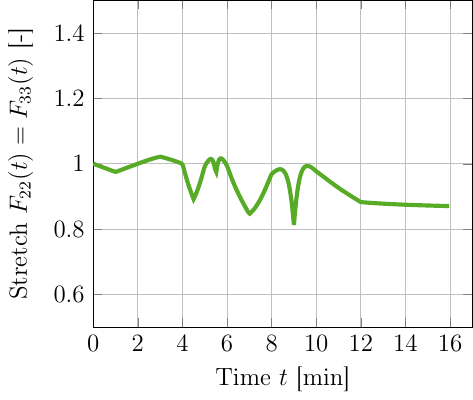}
    \caption{Components of the deformation gradient for uniaxial loading corresponding to the Helmholtz free energy~\eqref{eq:energy_artificial_1} and potential~\eqref{eq:potential_artificial_1} for the first example of artificial data.
    The loading path in the main direction is shown in Figure~\ref{fig:loading}. The remaining shear components of the deformation gradient are equal to zero.}
    \label{fig:results_artificial_1_deformation}
\end{figure}

In this first example, we investigate the network's capabilities to discover the weights and biases of our proposed network in order to accurately describe the material behavior.
As alluded in the previous section, we create an artificial data set by subjecting a constitutive model to a complex loading path.
For the Helmholtz free energy and the dual potential, we use the following compressible Neo-Hookean model
\begin{align}
    \psi &= \frac{\mu}{2}\,\left(\tilde{I}_1^{\bar{\bm{C}}_e} - 3 \right) + \frac{\kappa}{4}\, \left( I_3^{\bar{\bm{C}}_e} - 1 - \mathrm{ln}\left(I_3^{\bar{\bm{C}}_e}\right)\right) + \frac{\mu}{2}\,\left(\tilde{I}_1^{\bm{C}} - 3 \right) + \frac{\kappa}{4}\, \left( I_3^{\bm{C}} - 1 - \mathrm{ln}\left(I_3^{\bm{C}}\right)\right) \label{eq:energy_artificial_1}\\
    \omega^* &= K_1\, \left(\mathrm{cosh}\left( \frac{J_3^{\bar{\bm{\Sigma}}}}{J_2^{\bar{\bm{\Sigma}}} + 1} \right) - 1 \right) \label{eq:potential_artificial_1}
\end{align}
with $\mu=1$ [MPa] and $\kappa=1$ [MPa], and $K_1=2$ [1/MPa].
From a rheological viewpoint, Equations~\eqref{eq:energy_artificial_1} and \eqref{eq:potential_artificial_1} describe a `spring' element connected in parallel to a Maxwell element.
Remarkably, the potential is not `analytically' included in the architecture of the network, but is otherwise arbitrary.
However, since neural networks serve as universal approximators, we want to explore the geneeralized iCANN's capabilities to discover this potential.
Additionally, the material parameters are chosen such that there is a high amount of energy dissipation present.
The components of the deformation gradient for testing are shown in Figure~\ref{fig:results_artificial_1_deformation}.
\newline

\begin{figure}[h]
    \centering
    \includegraphics[]{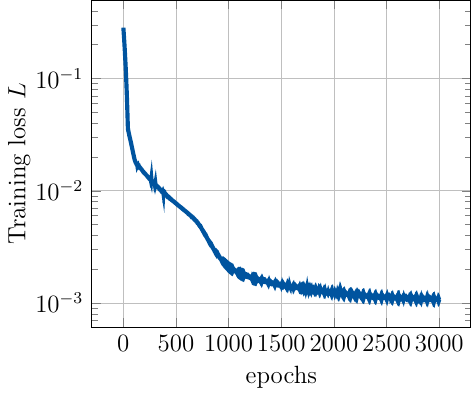}
    \caption{Loss during training of the generalized iCANN for the first example of artificially generated data. The loss is plotted on a logarithmic scale. For training, $3000$ epochs were used.}
    \label{fig:loss_artificial_1}
\end{figure}

\begin{figure}[h]
    \centering
    \includegraphics[]{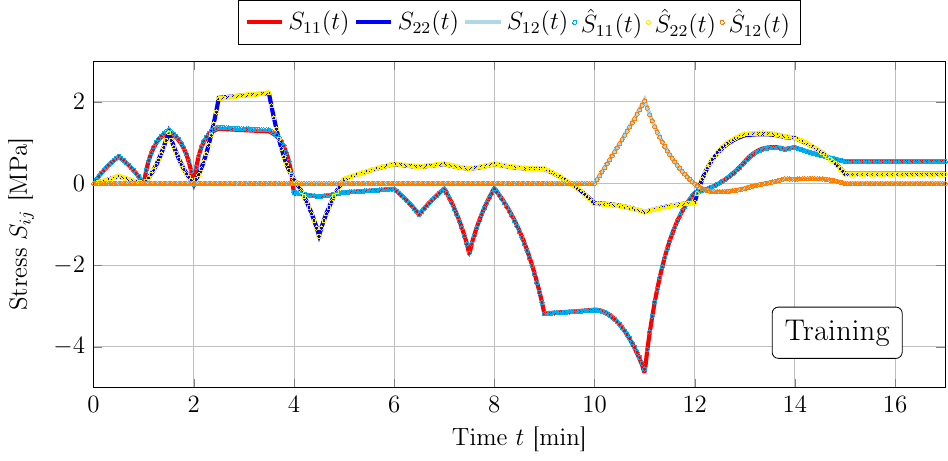}

    \includegraphics[]{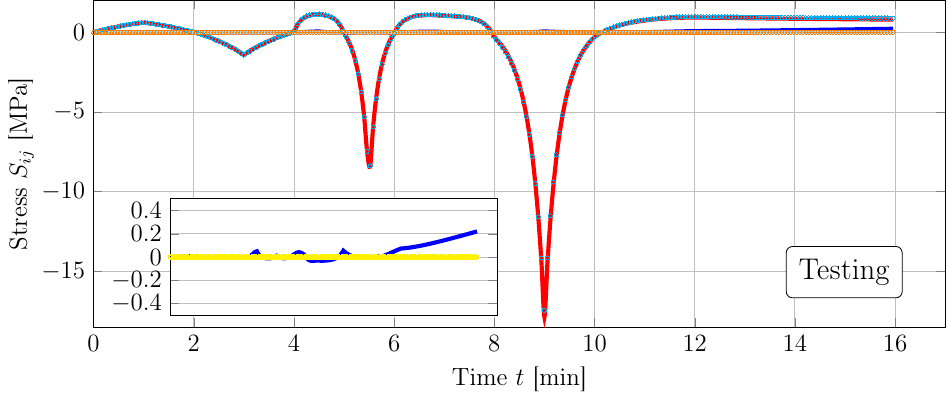}    
    \caption{Discovered model for the first example of artificially generated data. The predicted stress components, $S_{ij}$, are plotted against the artificially generated data, $\hat{S}_{ij}$. Above: Training results for the discovered weights of multiaxial loading. Below: Testing results for uniaxial tension. The zoomed window shows stresses $S_{22}$ and $\hat{S}_{22}$ over the entire load path up to $t=16$ [min].}
    \label{fig:train_test_artificial_1}
\end{figure}

\textbf{Discovery.} The weights identified in this study are too numerous to enumerate in this manuscript\footnote{Comprehensive data on weights and biases is available online.}. 
However, key components of the weight vector, $\mathbf{w}_{\omega^*}$, are summarized in Table~\ref{tab:w_pot_artificial_1}, with the corresponding training loss illustrated in Figure~\ref{fig:loss_artificial_1}. 
Notably, during training, the iCANN autonomously identifies the components of the first Maxwell element, $\prescript{1}{}{}\mathbf{w}_{\omega^*}$, as a zero vector, while the second potential exhibits relative sparsity. 
This finding is significant as it aligns with the rheological model defined by Equations~\eqref{eq:energy_artificial_1}-\eqref{eq:potential_artificial_1}. 
If the weights associated with the final layer of the generalized potential are zero, it indicates that the iCANN effectively simplifies to a Constitutive Artificial Neural Network (CANN) focused solely on hyperelastic behavior.
Figure~\ref{fig:train_test_artificial_1} presents the training and testing results for the weights identified during training. 
We observe good agreement between the artificial data and the model's response in both training and testing phases.
The model is capable of adequately recovering the relaxation behavior, especially in the constant deformation states, e.g., from $t=2.5$ [min] up to $t=3.5$ [min].

Although the test results align well in the primary loading direction, \(\hat{S}_{11}\), an artificial stress increase is observed in the off-axis direction, \(\hat{S}_{22}\). 
This suggests that the discovered weights deviate from the exact solution. If they perfectly matched the artificial material model, \(\hat{S}_{22}\) should vanish under uniaxial tension (Figures~\ref{fig:loading} and \ref{fig:results_artificial_1_deformation}). 
Instead, the discovered weights induce a constrained deformation, analogous to the difference between Young's modulus and the longitudinal modulus in elasticity theory.
Whether this truly constitutes an `issue' is debatable. 
If the discovered network is used to simulate a boundary value problem, the strong form of equilibrium would enforce a uniaxial stress state, albeit with a different transverse elongation.
\newline

\textbf{Implicit discovery.} Up to now, we have determined the network's weights and biases using an explicit exponential integrator scheme. 
However, implicit time integration schemes are more commonly employed in computational mechanics due to their superior stability and robustness. 

\begin{figure}[h]
    \centering
    \includegraphics[]{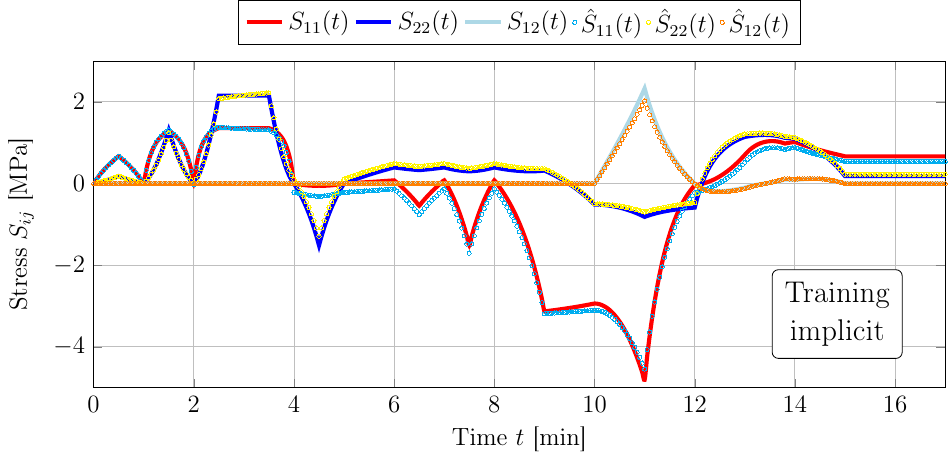}

    \includegraphics[]{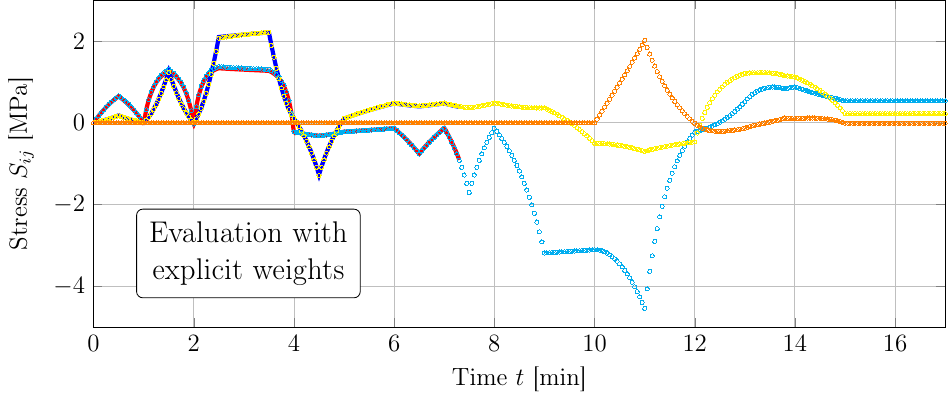}
    \caption{\textbf{Implicit discovery.} Above: Discovered model for the first example of artificially generated data with an implicit time integration scheme. The network's weights for the generalized dual potential, $\prescript{1}{}{}\mathbf{w}_{\omega^*}$ and $\prescript{2}{}{}\mathbf{w}_{\omega^*}$, turn out to be the zero vector. Thus, the network behaves as a purely elastic network (CANN). Below: Predicted stresses for the generalized iCANN with implicit time integration but employing the discovered weights and biases of the explicit architecture (cf. Figure~\ref{fig:train_test_artificial_1}). The local iteration aborts after $t=7.35$ [min].}
    \label{fig:implicit_artificial_1}
\end{figure}

To investigate this, we trained the implicit version of the generalized iCANN using the same data set. 
The results, shown in Figure~\ref{fig:implicit_artificial_1}, reveal that both weight vectors, \(\prescript{1}{}{}\mathbf{w}_{\omega^*} = \prescript{2}{}{}\mathbf{w}_{\omega^*} = \mathbf{0}\), reduce to the zero vector. 
This implies that the iCANN collapses entirely into a CANN, rendering it incapable of representing inelastic material behavior. 
Consequently, we observe a strong discrepancy in the training results.

To further analyze this behavior, we evaluate the implicit network using the weights obtained from the explicit scheme (see Figure~\ref{fig:train_test_artificial_1}). 
Up to \(t = 7.35\) [min], the results align well with the training data. 
However, beyond this point, the local Broyden iteration fails to converge. 
This suggests that greater attention must be given to developing a stable and robust iterative solving technique, though this is beyond the scope of this study. 
Nevertheless, Section~\ref{sec:discussion} provides a more detailed discussion on the implementation of the implicit integration scheme.
\begin{table}[t]
    \centering
    %\resizebox{\textwidth}{!}{
        \begin{tabular}{l | l l l l l l l l}
                &   1   &   2   &   3   &   4   &   5   &   6   &   7   &   8 \\
            \hline
            $\prescript{1}{}{}\mathbf{w}_{\omega^*}$ & $0$ & $0$ & $0$ & $0$ & $0$ & $0$ & $0$ & $0$ \\
            $\prescript{2}{}{}\mathbf{w}_{\omega^*}$ & $0$ & $0.2103749$ & $0$ & $0$ & $0$ & $0$ & $0$ & $0$
        \end{tabular}
    %}
    \caption{First example of artificially generated data. Listed are the components of the weight vector, $\mathbf{w}_{\omega^*}$, for the generalized iCANN shown in Figure~\ref{fig:rheo}. The first four weights correspond to the maximum activation function, while the latter four correspond to the exponential activation.}
    \label{tab:w_pot_artificial_1}
\end{table}
\FloatBarrier
%================================================
\subsubsection{Discovering a model for artificial data: Example 2}
\label{sec:results_artificial_2}
\begin{figure}[h]
    \centering
    \includegraphics[]{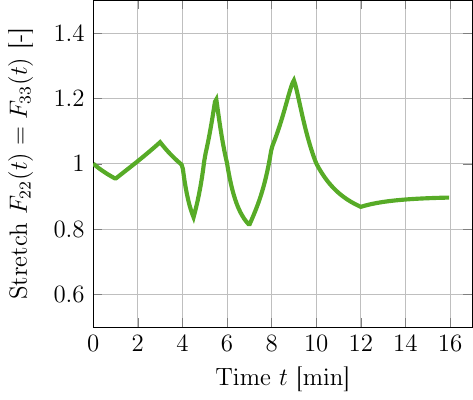}
    \caption{Components of the deformation gradient for uniaxial loading corresponding to the Helmholtz free energy~\eqref{eq:energy_artificial_1} and potential~\eqref{eq:potential_artificial_2} for the second example of artificial data.
    The loading path in the main direction is shown in Figure~\ref{fig:loading}. The remaining shear components of the deformation gradient are equal to zero.}
    \label{fig:results_artificial_2_deformation}
\end{figure}

We proceed by evaluating the performance of the generalized iCANN using a different constitutive model.
For the Helmholtz free energy, we stick to the very same energy~\eqref{eq:energy_artificial_1} as in the first example, however, we exchange the potential~\eqref{eq:potential_artificial_1} by a quadratic form, i.e.,
\begin{equation}
    \omega^* = K_1\, \left(2\,J_2^{\bar{\bm{\Sigma}}}\right) + K_2\, \left(I_2^{\bar{\bm{\Sigma}}}\right)^2
\label{eq:potential_artificial_2}
\end{equation}
with $\mu=25$ [MPa], $\kappa=50$ [MPa], $K_1=4 \cdot 10^{-5}$ [1/MPa], and $K_2=7.2 \cdot 10^{-4}$ [1/MPa].
The shear modulus, $\mu$, as well as the bulk modulus, $\kappa$, are also used to determine the Helmholtz free energy~\eqref{eq:energy_artificial_1}.
As in the first example, the constitutive model represents a Maxwell and a `spring' element connected in parallel.
However, contrary to the first example, the choice of material parameters yields a small amount of energy dissipation compared to the elastic response of the `spring'.
To create artificial data sets for training and testing, we subject the constitutive model (Equations~\eqref{eq:energy_artificial_1} and \eqref{eq:potential_artificial_2}) to the loadings described in Figure~\ref{fig:loading}.
The remaining components of the deformation gradient to achieve a state of uniaxial tension are shown in Figure~\ref{fig:results_artificial_2_deformation}.\newline

\textbf{Discovery.} For the chosen material parameters, the training stresses exceed $100$ [MPa], causing the iteration to abort. 
To address this, we normalize the artificial data by the absolute maximum stress value, \( S^{\text{max}} = 125.1517 \) [MPa], as is common in neural network training.

\begin{figure}[h]
    \centering
    \includegraphics[]{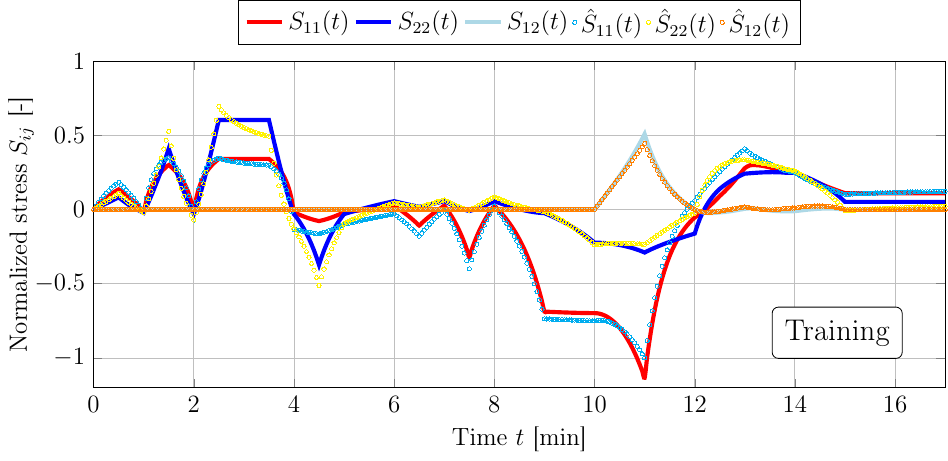}
    \caption{\textbf{Full data discovery.} Discovered model for the second example of artificially generated data. The predicted stress components, \( S_{ij} \), are plotted against the normalized artificially generated data, \( \hat{S}_{ij} \). The data are normalized by \( S^{\text{max}} = 125.1517 \) [MPa]. The entire load path, consisting of 341 data points, is used simultaneously for training.}
    \label{fig:train_full_artificial_2}
\end{figure}

Figure~\ref{fig:train_full_artificial_2} presents the training results. 
As in the first example, the entire load path up to \( t = 17 \) [min], consisting of 341 data points, is used for training. 
Once again, the iCANN autonomously identifies that the weights of one Maxwell element vanish while the other retain nonzero values. 
However, the predicted stresses exhibit poor agreement with the artificial data, particularly in capturing the relaxation behavior.

This discrepancy raises the question of its origin. 
Given that the same complex loading path is used, the issue is unlikely due to insufficient data richness. 
A potential bottleneck, particularly for inelastic materials, may be the initialization of the weights.\newline 

\textbf{Staggered discovery.} To investigate our assumption regarding initialization, we vary the batch size used for training, i.e., we do not utilize the full data set at once. 
Instead, we start with a smaller subset of the 341 data points. 

Although varying the batch size is a common practice in neural network training and can be applied to CANNs, inelastic materials require special treatment. 
Due to the strong sequential dependency imposed by the time integration scheme, randomly selecting data points is not feasible. 
Instead, each training batch must consist of a continuous sequence of data points.
Furthermore, even when using sequential subsets, the initial values of the internal states remain unknown. 
Consequently, every batch must begin at the very start of the entire load path to ensure consistency in training.

\begin{figure}[h]
    \centering
    \includegraphics[]{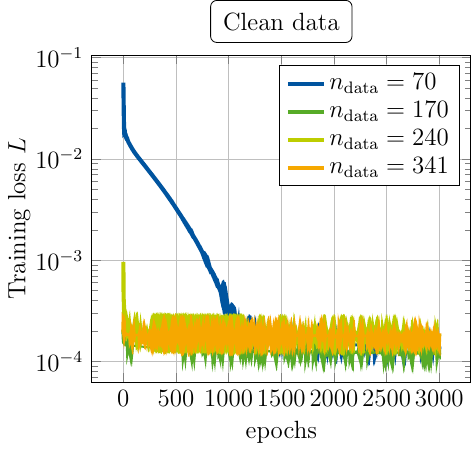}
    \includegraphics[]{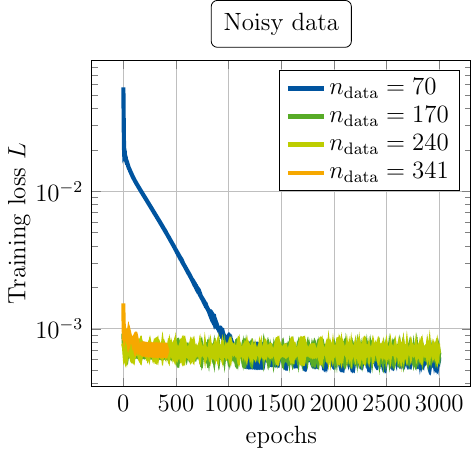}
    \caption{Losses during training of the generalized iCANN for the second example of artificially generated data. The losses are plotted on a logarithmic scale. For training, $3000$ epochs were used. A staggered discovery scheme is employed, i.e., the data points used for training are gradually increased. Left: Losses for the `clean' data set, see Figure~\ref{fig:train_test_artificial_2}. Right: Losses for the `noisy' data set, see Figure~\ref{fig:train_test_artificial_noisy_2}.}
    \label{fig:loss_artificial_2}
\end{figure}

\begin{figure}[h]
    \centering
    \includegraphics[]{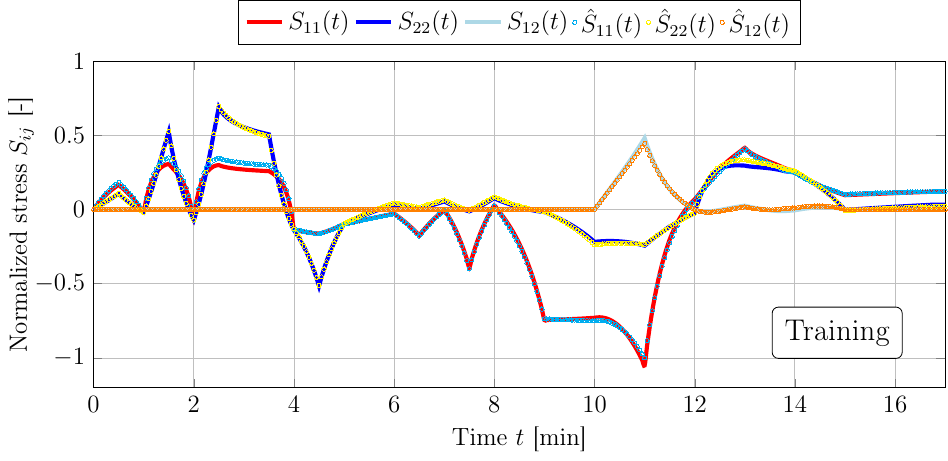}

    \includegraphics[]{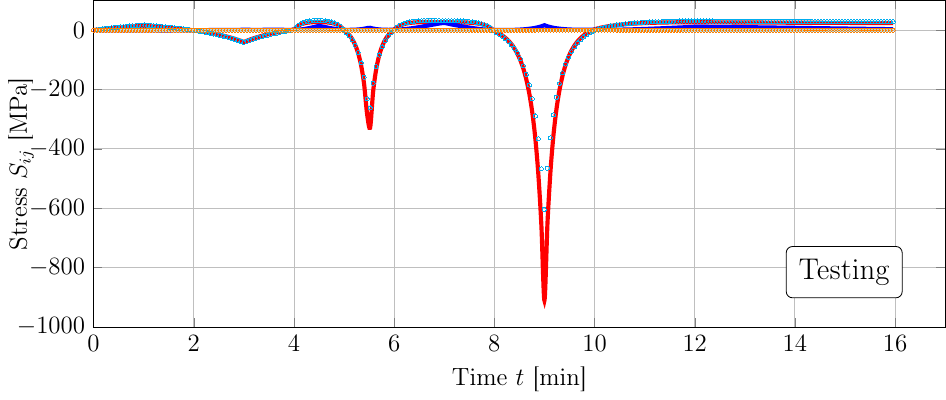}
    \caption{\textbf{Staggered discovery with clean data.} Discovered model for the second example of artificially generated data. The predicted stress components, $S_{ij}$, are plotted against the artificially generated data, $\hat{S}_{ij}$. Above: Training results for the discovered weights of multiaxial loading. The stresses are normalized by \( S^{\text{max}} = 125.1517 \) [MPa]. Below: Testing results for uniaxial tension. The stresses are not normalized.}
    \label{fig:train_test_artificial_2}
\end{figure}

In this contribution, we introduce a staggered discovery scheme, where training begins with a small subset of data points. 
The subset is then incrementally expanded while initializing each new training phase with the discovered weights from the previous step. 
This process is repeated until the entire load path is used for training.
For the current study, we employ three subsets: 70 data points, 170 data points, and 240 data points. 
The training losses for all sessions are plotted in Figure~\ref{fig:loss_artificial_2}, while the discovered weights of the last layer of the generalized potential are listed in Table~\ref{tab:w_pot_artificial_2}. 
Figure~\ref{fig:train_test_artificial_2} presents both the training and testing results.

It is noteworthy that the staggered discovery scheme yields better agreement with the artificial data, particularly in capturing the relaxation behavior, compared to the initial approach (cf. Figure~\ref{fig:train_full_artificial_2}). 
Once again, the iCANN autonomously identifies the visco-elastic nature of the data, as one iCANN reduces to a CANN network.
Although the potential is explicitly incorporated into the proposed network architecture, some discrepancies remain, indicating that the network has not yet discovered the optimal solution. 
Interestingly, despite the inclusion of the quadratic potential, the discovered weights in Table~\ref{tab:w_pot_artificial_2} reveal that only the weights associated with the exponential activation function are nonzero.

During testing, the predicted responses align well with the artificial data for most of the loading path, except at the second peak stress. 
This discrepancy may be attributed to the exponential activation function, which grows rapidly with increasing stress. 
Consequently, its incorrect influence may manifest only at high stress levels and was likely negligible during training.
As in the first example, an artificial increase in off-axis stresses is observed throughout the testing session. 
The underlying cause of this phenomenon is similar to the explanation provided in Section~\ref{sec:results_artificial_2}.

In this study, we did not adjust hyperparameters such as the number of layers, neurons, or activation functions, nor did we refine weight initialization. 
These factors may influence our findings (see Section~\ref{sec:discussion} for further discussion).\newline

\begin{figure}[h]
    \centering
    \includegraphics[]{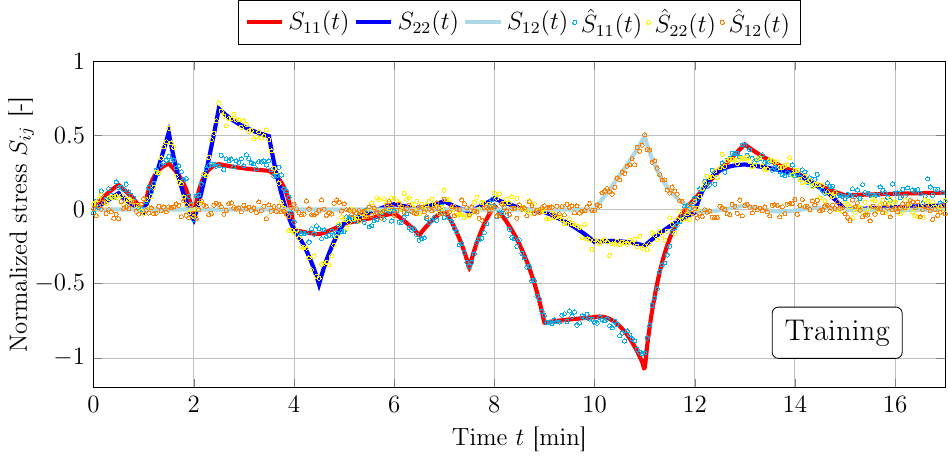}
    \caption{\textbf{Staggered discovery with noisy data.} Discovered model for the second example of artificially generated data. The predicted stress components, $S_{ij}$, are plotted against the artificially generated data, $\hat{S}_{ij}$. White Gaussian noise is added to the data shown in Figure~\ref{fig:train_test_artificial_2}.}
    \label{fig:train_test_artificial_noisy_2}
\end{figure}

\textbf{Discover noisy data.} So far, we have considered `clean' data generated deterministically from the constitutive model. 
However, real measurement data often contain noise. 
To evaluate the stability of our chosen regularization technique against overfitting, we introduce white Gaussian noise into our training data. 
We apply the staggered discovery scheme once again. 
The training results are presented in Figure~\ref{fig:train_test_artificial_noisy_2}, while losses are plotted in Figure~\ref{fig:loss_artificial_2}. 
Table~\ref{tab:w_pot_artificial_2} lists the weights discovered in the last layer of the generalized iCANN. 
The discovered model aligns well with the noisy data, showing no signs of overfitting; additionally, its qualitative response resembles that without noise. 
Notably, the same neuron is activated. 
However, it can be assumed that stress predictions may diverge from artificial data for higher stress peaks due to greater variability in inelastic stretch evolution.
\begin{table}[h]
    \centering
    %\resizebox{\textwidth}{!}{
        \begin{tabular}{l l | l l l l l l l l}
                &    &   1   &   2   &   3   &   4   &   5   &   6   &   7   &   8 \\
            \hline
            \multirow{2}{*}{Clean}  & $\prescript{1}{}{}\mathbf{w}_{\omega^*}$ & $0$ & $0$ & $0$ & $0$ & $0.00452437$ & $0$ & $0$ & $0$ \\
                                    & $\prescript{2}{}{}\mathbf{w}_{\omega^*}$ & $0$ & $0$ & $0$ & $0$ & $0$ & $0$ & $0$ & $0$ \\
            \hline
            \multirow{2}{*}{Noisy}  & $\prescript{1}{}{}\mathbf{w}_{\omega^*}$ & $0$ & $0$ & $0$ & $0$ & $0.086715$ & $0$ & $0$ & $0$ \\
                                    & $\prescript{2}{}{}\mathbf{w}_{\omega^*}$ & $0$ & $0$ & $0$ & $0$ & $0$ & $0$ & $0$ & $0$                                    
        \end{tabular}
    %}
    \caption{Second example of artificially generated data. Listed are the components of the weight vector, $\mathbf{w}_{\omega^*}$, for the generalized iCANN shown in Figure~\ref{fig:rheo}. The first four weights correspond to the maximum activation function, while the latter four correspond to the exponential activation. Both the weights for the clean data set (see Figure~\ref{fig:train_test_artificial_2}) and the noisy data set (see Figure~\ref{fig:train_test_artificial_noisy_2}) are shown.}
    \label{tab:w_pot_artificial_2}
\end{table}
\FloatBarrier
%================================================
\subsection{Discovering a model for the experimental data of VHB 4910 polymer}
\label{sec:Hossain2012}
In this final example, we investigate the generalized iCANN's ability to model experimentally measured data. 
We utilize experimental data from very-high-bond (VHB) 4910 polymer published by \citet{hossain2012}, which has previously been used to assess the original iCANN formulation \cite{holthusen2024} and has been compared with other constitutive neural network approaches \cite{AbdolaziziLinkaEtAl2023} as well as a classical constitutive model \cite{hossain2012}. 
In these comparisons, the iCANN demonstrated strong performance and effectively predicted outside the training regime. 
Thus, this example also facilitates a comparison between our newly proposed iCANN and existing methodologies.

The polymer is subjected to constant uniaxial loading followed by uniaxial unloading at three distinct but constant rates. 
Several experiments were conducted with varying maximum stretch levels, specifically $ C_{11}^{\text{max}} \in \{2.25, 4.0, 6.25, 9.0\} $ [-]. 
Notably, no experimental data is provided for a deformation rate of $ 0.03 $ [1/s] at the maximum stretch level applied. 
In contrast to our previous examples, the time increment in this case is not constant, and we normalize the stress data to its maximum value of $ S^{\text{max}} = 27.988124 $ [kPa].

The data is one-dimensional; thus, no information on transverse elongation is provided. 
This results in an undetermined problem, leading us to assume incompressibility in line with prior studies \cite{holthusen2024,holthusen2024PAMM,AbdolaziziLinkaEtAl2023}. 
For details on implementing incompressibility, interested readers are referred to the original iCANN work \cite{holthusen2024}.\newline

\textbf{Discovery: Training A.} We begin with the same training setup as in \cite{holthusen2024}, utilizing only the data for $ C_{11}^{\text{max}} = 9.0 $ [-] for training. 
The corresponding loss during training is depicted in Figure~\ref{fig:Hossain2012_losses}, while the weights of the last layer of the potential are presented in Table~\ref{tab:w_pot_Hossain2012}. 
Notably, we do not use the deformation rate of $ 0.03 $ [1/s] for training, leaving this rate unseen by the network. 
Furthermore, for the experimental data, the iCANN autonomously discovers visco-elasticity by reducing one iCANN to a CANN architecture, resulting in zero weights.

In contrast to identifying material parameters for classical constitutive models \cite{hossain2012}, we omit multi-step relaxation data from the training. 
Our training and testing results are shown in Figure~\ref{fig:Hossain2012_training_A}. 
While there is good agreement during training, the model does not perform well on testing data; few predictions match well, e.g., for $ C_{11}^{\text{max}} = 4.0 $ [-] with a constant deformation rate of $0.05$, while most do not.

A plausible explanation has been provided by \citet{AbdolaziziLinkaEtAl2023}: 
The experimental data exhibit inconsistencies; for identical loading rates but different maximum stretch levels, the loading paths should align, which is not observed (see \cite{holthusen2024}). 
Consequently, a neural network based on constitutive equations -- including our generalized iCANN -- struggles to predict such stress curves.

Despite our new proposed iCANN performing well during training, it is less accurate compared to previous works \cite{hossain2012,AbdolaziziLinkaEtAl2023} and particularly when compared to the original iCANN \cite{holthusen2024}. 
This may seem surprising at first, as the original architecture is included within the new one.

However, while the original architecture consists of only nine weights per Maxwell element for the feed-forward network of the potential, our new architecture employs 316 weights (see Section~\ref{sec:architecture_potential}). 
Thus, it is not surprising that one-dimensional data may lack sufficient complexity to accurately determine all network weights. 
Considering Shannon's information theory \cite{shannon1948} alongside our results from artificially generated data sets with richer information could provide insight into this issue (see also Section~\ref{sec:discussion}).

Lastly, we note that four Maxwell elements were used in the original iCANN architecture to adequately recover stress responses and that it was initialized with a `spring' (CANN) element.
In contrast, our newly proposed architecture autonomously reduces complexity and requires only a single Maxwell element within the generalized iCANN framework to accurately recover stress responses during training.\newline

\textbf{Discovery: Training B.} The results from Training A indicate the need for additional data to enhance model discovery. 
Therefore, we incorporate not only the data for $ C_{11}^{\text{max}} = 9.0 $ [-] but also that for $ C_{11}^{\text{max}} = 4.0 $ [-], including all three deformation rates in the training process.

\begin{figure}[h]
    \centering
    \hspace*{1cm}
    \begin{tikzpicture}[]
    \begin{axis}[xlabel= Nice $x$ label, ylabel= Nice $y$ label, width=55mm, height=35mm,
				hide axis,
				xmin=10,
				xmax=50,
				ymin=0,
				ymax=0.4,
				legend columns=-1,
				legend style={column sep=1mm}
				]
\addlegendimage{red, smooth,
        line width=2pt}
\addlegendentry{$\dot{F}_{11}=0.05$}
\addlegendimage{blue, 
        line width=2pt}
\addlegendentry{$\dot{F}_{11}=0.03$} 
\addlegendimage{lightblue, 
        line width=2pt}
\addlegendentry{$\dot{F}_{11}=0.01$} 
\addlegendimage{cyan, 
        mark=o,
        mark size=1.0pt,
        only marks,
        line width=0pt}
\addlegendentry{$\dot{\hat{F}}_{11}=0.05$}
\addlegendimage{yellow, 
        mark=o,
        mark size=1.0pt,
        only marks,
        line width=0pt}
\addlegendentry{$\dot{\hat{F}}_{11}=0.03$} 
\addlegendimage{orange, 
        mark=o,
        mark size=1.0pt,
        only marks,
        line width=0pt}
\addlegendentry{$\dot{\hat{F}}_{11}=0.01$ [1/s]} 
    \end{axis}
\end{tikzpicture}
    \vspace*{-1.2cm}
    
    \includegraphics[]{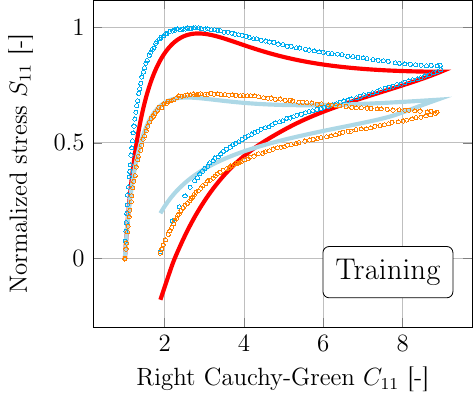}
    \includegraphics[]{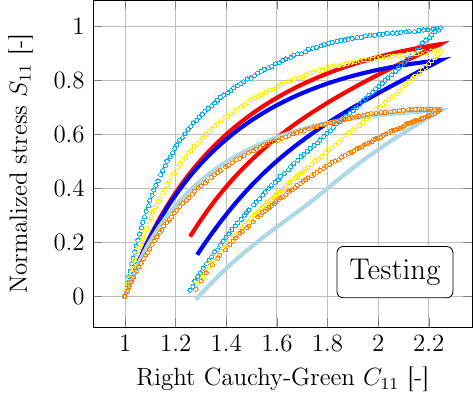}

    \includegraphics[]{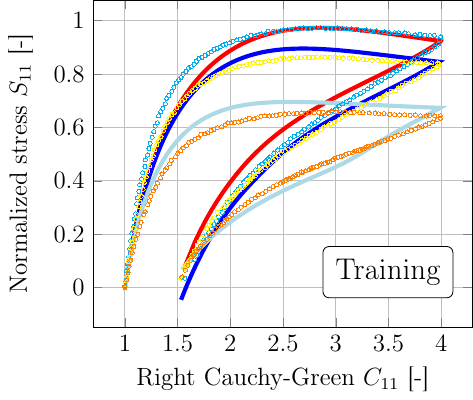}
    \includegraphics[]{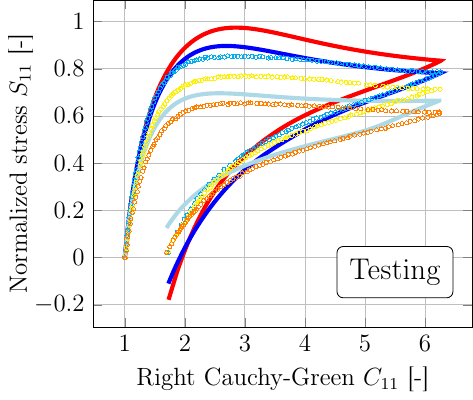}    
    \caption{\textbf{Training B.} Discovered model for the experimental data of VHB 4910 polymer \cite{hossain2012} employing three different constant loading/unloading rates, denoted as $\dot{F}_{11}$, for training. The experimentally measured data is represented by $\dot{\hat{F}}_{11}$. Stresses are normalized by $ S^{\text{max}} = 27.988124 $ [kPa]. The experimental data corresponding to $ C_{11}^{\text{max}} = 9.0 $ and $ C_{11}^{\text{max}} = 4.0 $ [-] are utilized for training.}
    \label{fig:Hossain2012_training_B}
\end{figure}

The loss is illustrated in Figure~\ref{fig:Hossain2012_losses}, and the discovered weights are detailed in Table~\ref{tab:w_pot_Hossain2012}. 
Once again, the iCANN autonomously identifies the underlying mechanisms of visco-elasticity by reducing one iCANN to a CANN. 
Our training and testing results are presented in Figure~\ref{fig:Hossain2012_training_B}.

Although performance improves within the training regime due to this richer dataset compared to Figure~\ref{fig:Hossain2012_training_A}, predictions outside this regime remain poor relative to earlier studies. 
While qualitative stress responses may be recovered, quantitative curves do not align well. 
The iCANN continues to struggle with insufficiently `rich' data for accurately determining weights and biases relevant to the material under investigation.

As previously noted, stresses should ideally match for identical deformation rates. 
Consequently, the additional data from $ C_{11}^{\text{max}} = 4.0 $ [-] primarily introduces previously unseen deformation rates and unloading curves. 
Whether this sufficiently enhances the richness of one-dimensional data for discovering numerous weights remains questionable (see also Section~\ref{sec:discussion}).
\begin{table}[t]
    \centering
    %\resizebox{\textwidth}{!}{
        \begin{tabular}{l l | l l l l l l l l}
                &    &   1   &   2   &   3   &   4   &   5   &   6   &   7   &   8 \\
            \hline
            \multirow{2}{*}{Training A}  & $\prescript{1}{}{}\mathbf{w}_{\omega^*}$ & $0$ & $0$ & $0$ & $0$ & $0.19749504$ & $0$ & $0$ & $0$ \\
                                    & $\prescript{2}{}{}\mathbf{w}_{\omega^*}$ & $0$ & $0$ & $0$ & $0$ & $0$ & $0$ & $0$ & $0$ \\
            \hline
            \multirow{2}{*}{Training B}  & $\prescript{1}{}{}\mathbf{w}_{\omega^*}$ & $0$ & $0$ & $0$ & $0$ & $0.09977169$ & $0$ & $0$ & $0$ \\
                                    & $\prescript{2}{}{}\mathbf{w}_{\omega^*}$ & $0$ & $0$ & $0$ & $0$ & $0$ & $0$ & $0$ & $0$                                    
        \end{tabular}
    %}
    \caption{Experimental data of VHB 4910 polymer \cite{hossain2012}. Listed are the components of the weight vector, $\mathbf{w}_{\omega^*}$, for the generalized iCANN shown in Figure~\ref{fig:rheo}. The first four weights correspond to the maximum activation function, while the latter four correspond to the exponential activation. Both the weights for the training session A (see Figure~\ref{fig:Hossain2012_training_A}) and training session B (see Figure~\ref{fig:Hossain2012_training_B}) are shown.}
    \label{tab:w_pot_Hossain2012}
\end{table}
\begin{figure}[h]
    \centering
    \begin{tikzpicture}
\begin{semilogyaxis} [grid = major,
    			xlabel = {epochs},
    			ylabel = {Training loss $L$},
    			width=0.45\textwidth,
    			height=0.4\textwidth,
    			/pgf/number format/1000 sep={},
    legend pos = north east,
    legend cell align={left}   
		]
    \addplot[rwth1, line width=2pt, smooth] table[x expr=\coordindex+1, y index=0] {graphs/Hossain2012_Fig42_losses.txt};
    \addplot[rwth5, line width=2pt, smooth] table[x expr=\coordindex+1, y index=0] {graphs/Hossain2012_Fig42_Fig32_losses.txt};
\legend{Training A,Training B}
\end{semilogyaxis}
\end{tikzpicture}
    \caption{Losses during training of the generalized iCANN for the experimental data of VHB 4910 polymer \cite{hossain2012}. The losses are plotted on a logarithmic scale. For training, $3000$ epochs were used. Two distinct training sessions are conducted: Training A (see Figure~\ref{fig:Hossain2012_training_A}), which utilizes a smaller data set, and Training B (see Figure~\ref{fig:Hossain2012_training_B}), which employs a larger data set.}
    \label{fig:Hossain2012_losses}
\end{figure}
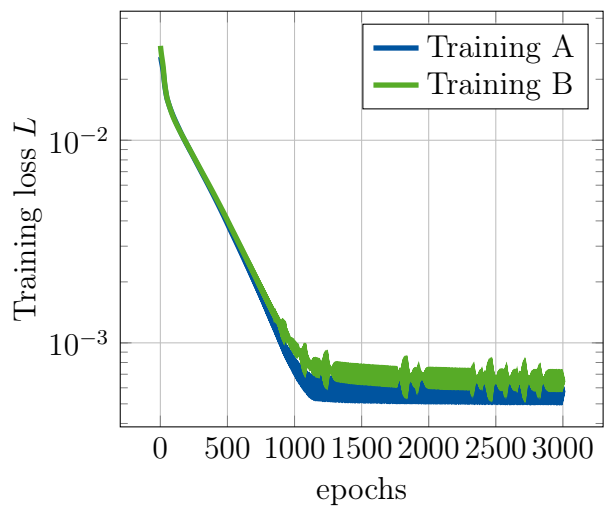
\begin{figure}[h]
    \centering
    \hspace*{1cm}
    \begin{tikzpicture}[]
    \begin{axis}[xlabel= Nice $x$ label, ylabel= Nice $y$ label, width=55mm, height=35mm,
				hide axis,
				xmin=10,
				xmax=50,
				ymin=0,
				ymax=0.4,
				legend columns=-1,
				legend style={column sep=1mm}
				]
\addlegendimage{red, smooth,
        line width=2pt}
\addlegendentry{$\dot{F}_{11}=0.05$}
\addlegendimage{blue, 
        line width=2pt}
\addlegendentry{$\dot{F}_{11}=0.03$} 
\addlegendimage{lightblue, 
        line width=2pt}
\addlegendentry{$\dot{F}_{11}=0.01$} 
\addlegendimage{cyan, 
        mark=o,
        mark size=1.0pt,
        only marks,
        line width=0pt}
\addlegendentry{$\dot{\hat{F}}_{11}=0.05$}
\addlegendimage{yellow, 
        mark=o,
        mark size=1.0pt,
        only marks,
        line width=0pt}
\addlegendentry{$\dot{\hat{F}}_{11}=0.03$} 
\addlegendimage{orange, 
        mark=o,
        mark size=1.0pt,
        only marks,
        line width=0pt}
\addlegendentry{$\dot{\hat{F}}_{11}=0.01$ [1/s]} 
    \end{axis}
\end{tikzpicture}
    \vspace*{-1.2cm}
    
    \includegraphics[]{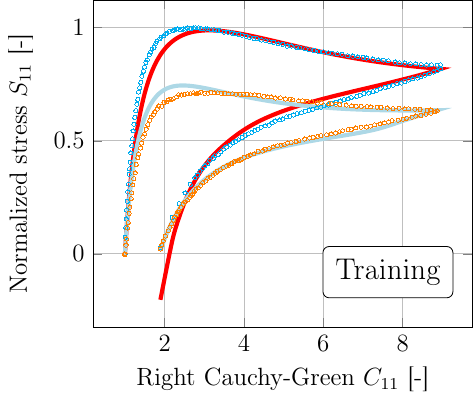}
    \includegraphics[]{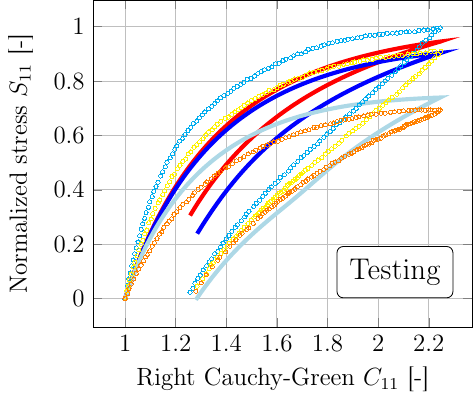}

    \includegraphics[]{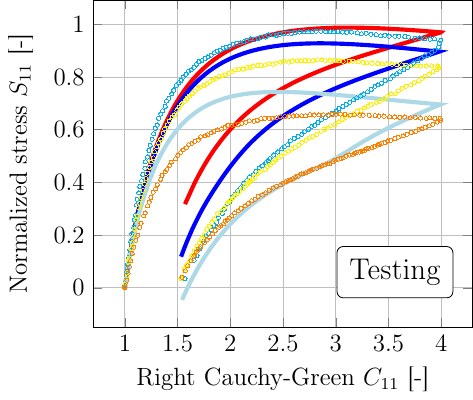}
    \includegraphics[]{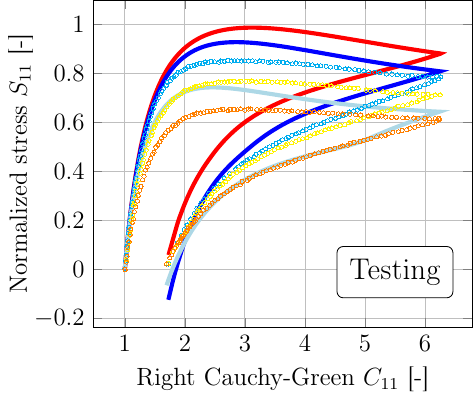}    
    \caption{\textbf{Training A.} Discovered model for the experimental data of VHB 4910 polymer \cite{hossain2012} employing three different constant loading/unloading rates, denoted as $\dot{F}_{11}$, for training. The experimentally measured data is represented by $\dot{\hat{F}}_{11}$. Stresses are normalized by $ S^{\text{max}} = 27.988124 $ [kPa]. Only the experimental data corresponding to $ C_{11}^{\text{max}} = 9.0 $ [-] are utilized for training.}
    \label{fig:Hossain2012_training_A}
\end{figure}
\FloatBarrier

\section{Discussion and current limitations}
\label{sec:discussion}
The results from the previous section demonstrated that the proposed iCANN architecture, whose complexity can be recursively increased along the lines of classical feed-forward networks, successfully discovers visco-elasticity at finite strains, and further, automatically uncovers the inelastic phenomenon hidden in the data. 
However, as with any approach that is still in its early stages, we faced some challenges during the discovery process, took some further steps towards effective discovery of inelastic materials, and were left with some unanswered questions, which we would like to share in the following.\newline

\textbf{Time integration scheme.} Our findings highlight challenges in the integration scheme. 
Training with an implicit scheme led to an undesired convergence of the weights of the potential towards zero.
Based on Section~\ref{sec:results_artificial_1}, we attribute this to inaccuracies in the iterative solver rather than the implicit integration itself. 
Broyden's method seems to be inadequate for solving the nonlinear evolution equation, requiring numerous local iterations (see \ref{app:broyden}), which complicates neural network backpropagation.

Initially, we employed Newton’s method, but automatic differentiation of the residual~\eqref{eq:implicit_integration} to get its Jacobian introduced issues in the backpropagation, prompting us to switch to the derivative-free Broyden method.

Further research is needed on combining neural network training with efficient iterative solvers to ensure stable training. 
The implicit function theorem, applied to the residual to calculate the derivative of the inelastic stretch with respect to the network's weights, may enhance stability. 
However, our studies show that explicit integration is computationally more efficient, though smaller time steps are required.
In real-world experiments, the step size can usually be controlled.\newline

\textbf{Regularization.} We implemented elastic net regularization for the weights connecting the inputs to the first hidden layer, while applying lasso regularization to the final layer. 
This decision aligns with empirical findings from CANNs. 
While our rationale for regularizing the last layer is somewhat justified, it remains a heuristic choice.

The use of elastic net is also primarily heuristic; however, it has demonstrated effective results and -- at least in our experience -- enhanced stability during training. 
It is important to note that heuristic strategies may not represent the optimal solution. 
The question of whether a superior regularization method exists for every inelastic phenomenon warrants further investigation.

In contrast to the Helmholtz free energy, we allow the potential to be zero. 
A zero energy implies no stresses would arise, which contradicts experimental data. 
A zero potential, on the other hand, leads to a hyperelastic model that could find an `optimal solution' for monotonic loading with constant stretch rates at different maximum stretch levels applied to the same material.

Interestingly, our chosen regularization approach resulted in sparse networks and appeared capable of identifying the degree of inelasticity within the data. 
Earlier training sessions conducted without regularization on the feed-forward network produced dense networks -- none of which reduced from iCANNs to CANNs, thereby failing to accurately represent visco-elastic solids.\newline

\textbf{Initialization.} We initialized the network weights using a uniform distribution, \( U(a,b) \), with heuristically determined bounds
\begin{equation}
\mathbf{W}_0^\tightI\sim U(-1,1),\quad 
\mathbf{W}_\tightI^\tightII,\,\mathbf{W}_\tightII^\tightIII\sim U(0,0.01),\quad 
\mathbf{w}_{\omega^*}\sim U(0,0.001),\quad 
p_1,\,p_2\sim U(0,1)
\end{equation}
Biases were initialized to zero to prevent symmetry-breaking effects.

Our training results indicate that performance is highly dependent on the amount of data used (see Section~\ref{sec:results_artificial_2}). 
When the data set is too large in combination with a low degree of inelasticity, the information entropy may become insufficient for discovering the optimal potential. 
Since we have not systematically explored alternative initialization bounds, we cannot exclude the possibility that different choices might enhance the network’s ability to generalize across the full data set.

Furthermore, in our experience, the ratio between the weights of energy and potential plays a crucial role. 
If energy weights are too large, the potential rapidly tends towards infinity regardless of its own weights. 
Conversely, if energy weights are too small, the potential may be significantly underestimated, leading to an overly elastic response. 
In extreme cases, the model may learn a purely elastic behavior that simply balances the data set rather than capturing the underlying material response.

Given the significant impact of weight initialization on learning dynamics, we recognize the need for systematic investigation. 
An evolutionary optimization approach, such as a genetic algorithm, could provide an effective strategy for weight distributions and mitigating undesirable energy-to-potential ratios.\newline

\textbf{Hyperparameters.} Across all examples, we employed the same set of hyperparameters, including learning rate, regularization parameters, number of hidden layers, number of neurons per activation function, choice of activation functions, and the clipping norm for gradient clipping. 
While some of these define the network architecture itself, such as depth and width, others regulate the training process.

Notably, model performance is influenced not only by the number of neurons, which likely correlates with the number of hidden layers, but also by training-specific parameters such as the clipping norm. 
The interplay among these factors renders hyperparameter selection a highly nonlinear optimization problem.

We observed that gradient clipping is essential for stabilizing the inelastic potential. 
Similar to weight initialization, where an improper energy-to-potential ratio can lead to an unbounded potential, the absence of gradient clipping can cause excessive growth of the potential’s weights between epochs, ultimately leading to divergence.

The chosen hyperparameters yielded satisfactory results for the examples considered in this study. 
However, we acknowledge that this may be due to a fortuitous selection of parameters and the specific network complexity in terms of depth and neuron count.

For artificial data, the training procedure performed well, producing results consistent with the data set. 
In contrast, performance deteriorated when applied to experimentally obtained data. 
This discrepancy may stem from the network's complexity, which introduces unnecessary flexibility in the learned weights. 
Furthermore, the information content of artificial and experimental data sets likely differs significantly.

Across all examples, the final state of training consistently produced a sparse weight vector \( \mathbf{w}_{\omega^*} \). 
Whether performance could be further improved through systematic hyperparameter tuning and reduced network complexity remains an open question.

We anticipate that hyperparameter optimization will be particularly critical for modeling more complex inelastic phenomena such as plasticity and damage.
Fortunately, hyperparameter optimization is an extensively studied problem in the neural network community. 
Established techniques such as reinforcement learning~\cite{wu2020}, grid search~\cite{bergstra2011}, genetic algorithms~\cite{sun2019}, and gradient-based optimization~\cite{pedregosa2016} offer promising strategies.
Recently, agent-based hyperparameter optimization \cite{esmaeili2023} has shown the potential to outperform these established techniques and may also serve as a valuable tool in physics-embedded neural networks.
For a comprehensive overview, we refer the reader to \citet{bischl2023}.\newline

\textbf{Richness of data sets.} Ultimately, the key question underlying our discussion is: 
How `rich' is the data? 
If the information entropy is sufficiently high, the iCANN architecture is likely capable of uncovering the inelastic phenomena hidden in the data and of accurately identifying the underlying potential governing the material’s response. 
However, the concept of `richness' in the context of material science requires further clarification.

Consider a previously discussed example: 
Suppose we conduct a series of uniaxial tension experiments on a visco-elastic solid, varying only the maximum stretch level while keeping the deformation rate constant across all tests.
Despite the large number of data points, the information entropy remains low. 
Consequently, the network will most likely identify a purely hyperelastic response. 
In contrast, if the deformation rate is also varied, the observed stress differences between experiments can only be attributed to the presence of an inelastic potential.

We encountered this issue with the VHB 4910 polymer (see Section~\ref{sec:Hossain2012}). 
While the network accurately reproduced the training data, its predictive performance was poor. 
This suggests that the `richness' of uniaxial tension data alone was insufficient to capture an accurate material model given the chosen network’s complexity. 
To illustrate this further, we consider an experiment in which a material undergoes hydrostatic relaxation, ensuring that no shear stresses arise. 
Since the second and third stress invariants remain zero throughout the experiment, the network would be unable to infer any material dependence on these invariants.

These challenges align with long-standing questions in experimental mechanics. 
Traditionally, experiments are designed to isolate specific material parameters, such as relaxation time in relaxation tests. 
However, the combination of neural networks and advanced optimization techniques -- both in training and hyperparameter tuning -- may enable us to replace numerous experiments with a single, highly complex experiment that captures comprehensive material behavior.

This raises an important question: 
How to uniquely design a `rich' experimental setup that captures not only elasticity and visco-elasticity but also phenomena such as plasticity, damage, and even multiphysics interactions?
We assume that leveraging structural discovery in complex boundary value problems will be instrumental in achieving this goal, as demonstrated by EUCLID~\cite{flaschel2022}.
%
%\FloatBarrier
\section{Conclusion}
\label{sec:conclusion}
We took a further step towards the understanding and discovery of general inelasticity at finite strains through a rigorous mathematical formulation that ensures a convex, non-negative, and zero-valued potential. 
This enabled the design of a scalable, interpretable network architecture akin to traditional feed-forward networks, encompassing various classical potentials in continuum mechanics. 
Our thermodynamically consistent approach inherently satisfies the dissipation inequality, ensuring predictions beyond training remain aligned with fundamental physical principles.

We introduced a regularization scheme inspired by Constitutive Artificial Neural Networks (CANNs), employing lasso and ridge techniques for sparse representation. 
This allowed the iCANN to autonomously determine the degree of inelasticity and reducing to a purely elastic CANN when appropriate. 
Gradient clipping prevented excessive weight growth in the potential network, avoiding unbounded stress responses and training instabilities.

Our studies successfully discovered multiple constitutive material models with high accuracy. 
A staggered discovery scheme prevented potential network weights from collapsing to zero when inelastic dissipation was low in the training data. 
The combination of our network architecture and regularization framework enabled stable identification of visco-elastic behavior, even with noisy data.

However, we encountered several limitations. 
The implicit time integration scheme was unstable when using Broyden's method to iteratively solve the evolution equation. 
Critically, while our network accurately captured training data for an experimentally measured polymer, it failed to predict accurately beyond the training regime, despite exploring various data set sizes. 
We attribute this to insufficient data richness, consistent with our findings for artificial data sets.

Balancing network complexity and data richness is a key challenge for meaningful weight discovery. 
The success of neural networks in approximating material behaviors hinges on sufficiently informative experimental data. 
This underscores the need for collaboration between experimental mechanics and computational modeling to design new experimental setups and discovery strategies. 
Future experiments may aim to maximize information content while minimizing test specimens and addressing uncertainties in material properties.

For our approach, this implies extending the network architecture to boundary value problems, leveraging displacement field information to enhance material discovery.

%%%%%% Main Text %%%%%%

\appendix
\section{Appendix}
%=========================================
\subsection{Broyden's method}
\label{app:broyden}
Algorithm~\ref{algo:broyden} presents Broyden's method \cite{broyden1965} to iteratively solve for the inelastic stretch, $\bm{U}_i$, in a derivative-free manner. 
To directly compute the inverse of the Jacobian, $\mathbf{B}$, the Sherman-Morrison formula \cite{sherman1950} is exploited.
\begin{algorithm}[h]
\caption{Broyden's method to determine $\bm{U}_i$ in Voigt's notation $(\bullet)'$}
\label{algo:broyden}
\begin{algorithmic}[1]
\State \textbf{Input:} tolerance \( \epsilon \), weights $\mathbf{W}$, biases $\mathbf{b}$, $\Delta t$, $\bm{C}_{n+1}$, $\bm{U}_{i_n}$
\State \textbf{Output:} $\bm{U}_i$
\State Set \( \bm{U}_i' \gets \bm{U}_{i_n}'\)
\State Initialize residual, $\mathbf{r}_0$, according to Equation~\eqref{eq:implicit_integration} in Voigt's notation
\State Set \( \mathbf{B} \gets \mathbf{I}_{6\times 6}\)
\For{$m = 1$ to $50$}
    \State $\mathbf{s} \gets -  \mathbf{B}\, \mathbf{r}_{m-1}$ 
    \State $\bm{U}_{i_m}' \gets \bm{U}_{i_{m-1}}' + \mathbf{s}$
    \State Compute new residual, $\mathbf{r}_m$, with updated inelastic stretch $\bm{U}_{i_m}$
    \State $\mathbf{y} \gets \mathbf{r}_m - \mathbf{r}_{m-1}$
    \State $\mathbf{B} \gets \mathbf{B} + \frac{\mathbf{s} - \mathbf{B}\,\mathbf{y}}{\mathbf{s}^T\mathbf{B}\,\mathbf{y}}\, \mathbf{s}^T\mathbf{B}$
    \If{$|\mathbf{r}_m| < \epsilon$}
        \State \textbf{break}
    \EndIf   
\EndFor
\State \textbf{return} $\bm{U}_i$
\end{algorithmic}
\end{algorithm}
\FloatBarrier
%
%=========================================
\subsection{Additional stress invariants}
\label{app:stress_invars}
We can find a relation between the additional stress invariants $I_2^{\bar{\bm{\Sigma}}}$ and $I_3^{\bar{\bm{\Sigma}}}$ and our three basic invariants as follows
\begin{align}
    I_2^{\bar{\bm{\Sigma}}} &= \frac{\left(I_1^{\bar{\bm{\Sigma}}}\right)^2}{6} + J_2^{\bar{\bm{\Sigma}}} \\
    I_3^{\bar{\bm{\Sigma}}} &= \frac{\left(I_1^{\bar{\bm{\Sigma}}}\right)^3}{27} + \frac{2}{3}\, I_1^{\bar{\bm{\Sigma}}}\, J_2^{\bar{\bm{\Sigma}}} + J_3^{\bar{\bm{\Sigma}}}.
\end{align}
While $I_2^{\bar{\bm{\Sigma}}}$ is included in the general architecture, we are unable to obtain $I_3^{\bar{\bm{\Sigma}}}$ since neither the activation function $(\bullet)^3$ nor the multiplication $I_1^{\bar{\bm{\Sigma}}}\, J_2^{\bar{\bm{\Sigma}}}$ is present in the general network architecture.
We can express the dual potential as $\omega^*=\tilde{\omega}^*\left(I_1^{\bar{\bm{\Sigma}}}, \sqrt{J_2^{\bar{\bm{\Sigma}}}}, \sqrt[3]{J_3^{\bar{\bm{\Sigma}}}}, \sqrt{I_2^{\bar{\bm{\Sigma}}}(I_1^{\bar{\bm{\Sigma}}},J_2^{\bar{\bm{\Sigma}}} )}, \sqrt[3]{I_3^{\bar{\bm{\Sigma}}}(I_1^{\bar{\bm{\Sigma}}},J_2^{\bar{\bm{\Sigma}}},J_3^{\bar{\bm{\Sigma}}} )} \right)$.
With these relations at hand, we find
\begin{align}
    \frac{\partial I_2^{\bar{\bm{\Sigma}}}}{\partial \bar{\bm{\Sigma}}} : \bar{\bm{\Sigma}} &= 2\,I_2^{\bar{\bm{\Sigma}}} \\
    \frac{\partial I_3^{\bar{\bm{\Sigma}}}}{\partial \bar{\bm{\Sigma}}} : \bar{\bm{\Sigma}} &= 3\,I_3^{\bar{\bm{\Sigma}}}
\end{align}
which, in analogy to Equation~\eqref{eq:reduced_dissipation_vector}, proves thermodynamic consistency if the specific network (Figure~\ref{fig:NN_potential}) is convex, zero-valued, and non-negative with respect to its five inputs.
\section{Declarations}
%=======================================================================
%
\subsection{Acknowledgements}
Hagen Holthusen and Tim Brepols gratefully acknowledge financial support of the projects 417002380 and 453596084 by the Deutsche Forschungsgemeinschaft.
In addition, Kevin Linka is supported by the Emmy Noether Grant 533187597 by the Deutsche Forschungsgemeinschaft.
This work was supported by the NSF CMMI Award 2320933 Automated Model Discovery for Soft Matter and by the ERC Advanced Grant 101141626 DISCOVER to Ellen Kuhl.
%
%=======================================================================
%
\subsection{Conflict of interest}
The authors of this work certify that they have no affiliations with or involvement in any organization or entity with any financial interest (such as honoraria; participation in speakers’ bureaus; membership, employment, consultancies, stock ownership, or other equity interest; and expert testimony or patent-licensing arrangements), or non-financial interest (such as personal or professional relationships, affiliations, knowledge or beliefs) in the subject matter or materials discussed in this manuscript.
%=======================================================================
%
\subsection{Availability of data and material}
Our data used for training and testing are accessible to the public at \url{https://doi.org/10.5281/zenodo.14894687}
%=======================================================================
%
\subsection{Code availability}
Our source code and examples of the iCANN implementation in JAX are accessible to the public at \url{https://doi.org/10.5281/zenodo.14894687}.
%=======================================================================
%
\subsection{Contributions by the authors}
\textbf{Hagen Holthusen:} Conceptualization, Methodology, Software, Validation, Formal analysis, Investigation, Data Curation, Writing - Original Draft, Writing - Review \& Editing, Visualization, Funding acquisition\\
\textbf{Kevin Linka:} Methodology, Writing - Original Draft, Writing - Review \& Editing, Funding acquisition\\
\textbf{Ellen Kuhl:} Methodology, Writing – Original draft, Writing – review \& editing, Funding acquisition\\
\textbf{Tim Brepols:} Methodology, Writing - Original Draft, Writing - Review \& Editing, Funding acquisition\\
%=======================================================================
%
\subsection{Statement of AI-assisted tools usage}
This document was prepared with the assistance of OpenAI's ChatGPT, an AI language model. ChatGPT was used for language refinement. The authors reviewed, edited, and take full responsibility for the content and conclusions of this work.

%\bibliography{literature}
%\printbibliography

\end{document}